\documentclass[10pt,conference]{IEEEtran}

\usepackage[T1]{fontenc}
\usepackage{textcomp}
\usepackage{cite}
\usepackage{times}
\usepackage{mathptmx}
\usepackage{amsmath,amssymb}
\usepackage{booktabs}
\usepackage{graphicx}
\usepackage[hidelinks]{hyperref}
\usepackage{url}

\begin{document}

\title{Uncovering the Blind Spots of OKS-mAP: A Unified Protocol for Occlusion, Joint-Angle, and Temporal Evaluation of 2D Pose Estimators}

\author{
\IEEEauthorblockN{Yifan Fan}
\IEEEauthorblockA{International Department, Taiyuan Yangfan School\\Taiyuan, China}
}

\maketitle

\begin{abstract}
The dominant evaluation protocol for 2D human pose estimation---mean average precision based on Object Keypoint Similarity (OKS-mAP)---is insensitive to three properties that matter for downstream applications: it does not distinguish visible from occluded keypoints in the aggregate score; it does not measure joint-angle error; and it is silent on the temporal stability of outputs produced by tracking-free, frame-by-frame inference. We propose a unified supplementary evaluation protocol with three modules: (M1) OKS-mAP stratified by ground-truth visibility labels ($v$), together with $\Delta_{\mathrm{occ}}=\mathrm{mAP}(v{=}1)-\mathrm{mAP}(\mathrm{full})$, which quantifies how much the aggregate metric masks occlusion-specific differences; (M2) joint-angle error (RMSE/MAE/P95) against angles derived from ground-truth keypoints, plus knee-valgus detection consistency (Cohen's $\kappa$); and (M3) frame-to-frame displacement, median-filter residual jitter, and jump rate under tracking-free per-frame inference, before and after OneEuro smoothing. We executed the protocol on four public datasets (COCO val2017, CrowdPose, OCHuman, PoseTrack18) for four models spanning two architecture families---the official YOLO11s-pose, an occlusion-domain fine-tuned variant, and the transformer-based ViTPose-B/L---with all cross-family comparisons performed under a common detector-free GT-bbox oracle protocol and all metrics produced by an open-source harness in a single inference pass. $\Delta_{\mathrm{occ}}$ is negative for every model--dataset combination ($-0.19$ to $-0.47$): the aggregate mAP overestimates occluded-scene performance in every combination tested, across both architecture families. Joint-angle error is amplified in occluded layers across all tested combinations, and on the occlusion-domain datasets the angle metrics reveal significant differences between models that the aggregate mAP cannot localize---including a case, now under the shared oracle protocol, where the smaller occlusion-domain-trained model is significantly more accurate in the occluded elbow layer than a larger transformer. On the matched retained-segment subpopulation of PoseTrack18 val (about 53\% of annotated frames), we found no evidence of a difference in raw temporal jitter between the two YOLO models under either protocol, while pixel-domain OneEuro smoothing reduces the two YOLO models' jump rates by 9--10\% at zero cost on the matched-scope AP ($|\Delta\mathrm{AP}|\le 0.0002$); the oracle protocol further shows the residual high-frequency jitter to persist after detector removal (it is not detector-intrinsic), with the transformer pair's jump rate significantly lower than the YOLO pair's under the shared protocol, and the smoothing-domain finding replicates across architectures. Correlation analysis shows that stratified mAP is highly consistent with full mAP (sign-consistency $\tau_s=1.0$, 12/12 pairs)---its value lies in magnitude diagnosis rather than re-ranking---while the angle metrics' difference vectors, particularly knee angle, do not fully align with the full-mAP ranking, suggesting they carry partially independent information (descriptive $n=6$, Section~5.4). The protocol and harness are open-sourced and positioned as a complement to, not a replacement for, OKS-mAP.
\end{abstract}

\begin{IEEEkeywords}
human pose estimation; evaluation protocol; occlusion; joint angle; temporal stability; OKS mAP
\end{IEEEkeywords}

\section{Introduction}

2D human pose estimation is a fundamental computer vision task whose outputs feed action recognition, motion analysis, and real-time fitness-correction systems. For such applications, the gap between benchmark scores and real-world usability matters more than average precision alone: a model that ranks first on a benchmark can be markedly worse in scenes with occluded people, fast motion, or the need for precise joint-angle judgments---and this difference is invisible in the dominant metric. Whether such blind spots change model rankings or only mask magnitudes is itself an empirical question; we test it (RQ1) and find, for the model pairs studied, that the ranking is stable within each architecture family while the magnitudes differ substantially (Sections~5.4 and 5.5).

The de facto standard evaluation protocol is OKS-mAP as defined by the COCO keypoints protocol~\cite{1}. It matches and scores predictions by Object Keypoint Similarity (OKS) and aggregates over the whole test set into a single scalar. This design has three blind spots. First, OKS aggregation ignores keypoint visibility: visible ($v{=}2$) and occluded ($v{=}1$) keypoints are averaged with equal weight into one score (missing, $v{=}0$, keypoints are excluded from OKS averaging), so occlusion-specific accuracy differences are diluted by averaging. Second, the protocol does not measure joint angles: OKS penalizes pixel-level keypoint position error, whereas downstream applications (e.g., exercise correction) care about whether the joint angles derived from those keypoints are accurate---and the two are not linearly related. Third, the protocol is blind to temporal behavior: most systems run frame-by-frame independent inference on video, and the jitter of the output between frames directly determines how the downstream consumer perceives it, yet a static-image protocol says nothing about it. Prior work responds to each problem in isolation---OCHuman organizes evaluation by occlusion level~\cite{2}; PoseTrack and SmoothNet extend the static protocol with tracking metrics (MOTA/MOTP) and a temporal-smoothness metric (per-joint acceleration magnitude, Accel), respectively~\cite{3,15}; and the biomechanics community has validated joint angles from markerless pose estimation~\cite{4,5,6}. What is missing is a unified, reproducible, cross-dataset supplementary protocol, and any baseline numbers quantifying how large these three blind spots are on real models.

This paper proposes a three-module unified supplementary evaluation protocol (Section~3) and open-sources its execution layer, an eval harness that produces official metrics and protocol metrics from a single inference pass, with three self-check gates. We executed the protocol on four public datasets with two representative models (Sections~4--5) to answer five research questions:

\begin{itemize}
\item \textbf{RQ1 (occlusion)}: How much does full OKS-mAP mask occlusion-specific accuracy differences? Does model ranking change after visibility stratification?
\item \textbf{RQ2 (angle)}: Does joint-angle RMSE provide discriminative information beyond OKS-mAP? Is angle error amplified by occlusion?
\item \textbf{RQ3 (temporal)}: Under tracking-free per-frame inference, how much do models differ in temporal stability? Can smoothing buy stability at an acceptable accuracy cost?
\item \textbf{RQ4 (protocol economy)}: How do the metrics of the three modules correlate with each other and with full mAP? Is any of them redundant?
\item \textbf{RQ5 (architecture generality)}: Under a detector-free GT-bbox oracle protocol, do the three blind spots persist for transformer-based models, and does model ranking re-order within another architecture family?
\end{itemize}

Our contributions are threefold: (1) to our knowledge, the first supplementary evaluation scheme that unifies occlusion stratification, angle error, and temporal jitter under one protocol, with explicit statistical conventions (bootstrap confidence intervals) and dataset-consistent stratification rules; (2) an open-source harness and all raw predictions, so that any new model can produce the full set of protocol metrics on the four datasets in one pass; (3) complete baseline numbers for four models from two architecture families---including a detector-free oracle protocol that isolates the pose head from the detector---that quantify, for the first time to our knowledge, the magnitude of OKS-mAP's three blind spots (e.g., $\Delta_{\mathrm{occ}}$ as large as $-0.47$ on COCO under the oracle protocol).

\section{Related Work}

\textbf{Pose estimation evaluation protocols.} The COCO keypoints protocol, with OKS at its core, has been the field's de facto standard since 2014~\cite{1}; its predecessor, percentage of correct keypoints (PCK) on MPII~\cite{16}, shares the same structure our modules target: one aggregate over the whole test set. PoseTrack extends evaluation to multi-person pose tracking in video with a tracking convention centered on MOTA/MOTP~\cite{3}; SmoothNet goes further and proposes per-joint acceleration magnitude (Accel, in pixels/frame$^2$) as a temporal-smoothness metric for video pose, and uses that convention in its filtering-vs-baseline comparisons~\cite{15}. Our M3 is complementary to Accel: Accel measures the size of the frame-to-frame second difference, whereas our frame-to-frame displacement and median-filter residual measure positional fluctuation and high-frequency noise separately, and we additionally report the accuracy (AP) cost of smoothing. On the data side, CrowdPose and OCHuman provide real visibility/occlusion labels ($v{=}1$ in CrowdPose; three occlusion levels in OCHuman)~\cite{2,7}. The common boundary of this prior work is that each improves a single dataset in a single dimension; none of it forms a cross-dataset supplementary protocol with unified conventions.

\textbf{Occlusion-robustness evaluation.} Prior work has attempted to quantify occlusion's effect on pose estimation: Ghafoor and Mahmood build three experiments with synthetically removed joints to quantify occlusion's impact on 3D pose error~\cite{8}, and later work continues this line with robustness evaluation under synthetic occlusion~\cite{9} and a visibility-aware video occlusion benchmark~\cite{10}. The difference from our work: these studies use synthetic occlusions or non-real annotation-based stratification, and each focuses on a single dataset and a single metric. We stratify across datasets using real $v$ labels from three datasets and report $\Delta_{\mathrm{occ}}$ against full mAP.

\textbf{Joint-angle validation.} Biomechanics and sports science have a line of validity studies for markerless pose estimation: Turner et al.\ reconstruct joint-angle trajectories from BlazePose keypoints in 89 resistance-training videos and assess movement-technique quality by frame-wise RMSE against a reference movement~\cite{4}; Panconi et al.\ validate DeepLabCut joint angles in gait analysis~\cite{5}; Pemasiri et al.\ build a gait-analysis framework for biomechanical interpretation~\cite{6}. These studies answer the question ``do estimated angles equal the clinical/marker-based gold standard?'' (validity), and they require dedicated marker or expert annotations. We do not repeat validity validation; instead we use ground-truth-derived angles as a protocol metric for model-to-model comparison. The two lines are complementary rather than competing (Section~6 elaborates this boundary). The knee-valgus threshold takes its magnitude reference from the literature: frontal plane projection angle (FPPA) of 16.0--17.4$^{\circ}$ in a lateral step-down test~\cite{11}.

\textbf{Temporal consistency and smoothing.} Temporal inconsistency of tracking-free per-frame inference has been discussed at a general level by Deep Video Prior and others~\cite{12}; in engineering practice, OneEuro causal filtering is widely used to smooth keypoint trajectories~\cite{13}. Our work turns ``stability of tracking-free raw output'' itself into a protocol metric, and compares two filtering domains (normalized-coordinate vs.\ pixel) in their joint effect on jitter metrics and AP---a comparison that reveals how the choice of filtering domain decides the benefit and cost of smoothing.

\section{Protocol Design}

\subsection{Overview}

The protocol consists of three modules plus one statistical layer, all built on the common foundation of the COCO 17-keypoint definition and the OKS convention. Table~\ref{tab:overview} gives the overview.

\begin{table}[t]
\centering
\caption{Protocol overview.}
\label{tab:overview}
\renewcommand{\arraystretch}{1.15}
\resizebox{\columnwidth}{!}{%
\begin{tabular}{p{0.95in}p{0.85in}p{1.15in}p{0.45in}}
\toprule
\textbf{Module} & \textbf{Metrics} & \textbf{Data} & \textbf{Answers} \\
\midrule
M1 occlusion stratification & $v$-stratified mAP$^{50-95}$(P) + $\Delta_{\mathrm{occ}}$ & CrowdPose val / OCHuman test / COCO val & RQ1 \\
M2 angle error & elbow/knee angle RMSE/MAE/P95 + knee-valgus detection consistency ($\kappa$) & same as M1 (angles derived from GT, zero extra annotation) & RQ2 \\
M3 temporal jitter & frame-to-frame displacement / jitter / jump rate, before and after OneEuro & PoseTrack18 val (tracking-free per-frame inference) & RQ3 \\
Statistical layer & 10k bootstrap CI + sign-consistency $\tau_s$ + Spearman $\rho$ & results of the three modules & RQ4, model comparison \\
\bottomrule
\end{tabular}}
\end{table}

\subsection{M1: Occlusion-Stratified mAP}\textbf{Stratification convention.} Each GT keypoint is assigned to a layer by its visibility label $v$: $v{=}2$ visible, $v{=}1$ occluded, $v{=}0$ missing. OKS AP is computed independently within each layer. The $v{=}0$ layer is reported but excluded from ranking (keypoints in this layer cannot be estimated and scores are near zero; the layer is kept as evidence of protocol completeness). Each dataset retains its real $v$ labels after conversion to COCO17: CrowdPose's official 14 keypoints map to COCO indices 5--16 (the head\_top and neck points are dropped---a mapping-induced truncation, not a corruption of visibility labels; the $v{=}1$ labels of the 12 mapped points are fully preserved); OCHuman's official test annotations are already in COCO17 format and pass through unchanged; COCO val itself contains $v{=}1$ labels (7.75\% of keypoints).

\textbf{Computation convention.} We follow the keypoints protocol of pycocotools 2.0.11 (the convention behind official released numbers): mean over 10 OKS thresholds 0.50:0.95; GTs with num\_keypoints$=$0 marked ignore; maxDets$=$20. Stratification changes only the statistics, not the matching logic: person matching (TP/FP decisions) still uses full-keypoint OKS, and a layer's AP accumulates only the OKS contributions of that layer's GT keypoints. Three gates guarantee implementation correctness: Gate 1 agrees with the official pycocotools 2.0.11 library within 0.0001; Gate 2 is a synthetic self-check ((i) a GT-derived-angle passthrough prediction path reproduces the target RMSE $<10^{-9}$; (ii) a binarized knee-valgus sequence identical to GT yields sensitivity $=$ specificity $=1$; (iii) constant trajectories have zero displacement and jitter; (iv) a constant sequence is unchanged by OneEuro); Gate 3 verifies that on all-$v{=}2$ synthetic data the $v{=}2$-layer AP equals the full AP and the stratification toggle leaves full mAP unchanged.

\textbf{Interpretability evidence.} A layer's AP has all GT persons as its recall denominator, so persons without that layer's keypoints can never contribute TPs to it. We therefore report, per layer, both ``fraction of GT keypoints in this layer'' (gt\_frac) and ``fraction of GT persons possessing this layer's keypoints'' (gt\_person\_frac). In particular, CrowdPose's official annotations mark nearly all visible points as $v{=}2$ (occluded points cluster at $v{=}1$), while a layer's AP has all GT persons as its recall denominator---gt\_person\_frac is only 0.591---so its v2-layer AP is systematically underestimated. This denominator bias acts on the v2 layer and on the full AP alike and cancels in the difference $\Delta_{\mathrm{occ}}=\mathrm{mAP}(v{=}1)-\mathrm{mAP}(\mathrm{full})$---which is exactly why cross-layer comparisons on CrowdPose use $\Delta_{\mathrm{occ}}$.

\textbf{Primary metric.} $\Delta_{\mathrm{occ}}=\mathrm{mAP}(v{=}1\ \mathrm{layer})-\mathrm{mAP}(\mathrm{full})$; a negative value means the full metric overestimates occluded-scene performance. Both terms are computed over the same image universe and the same full GT person set (the $v{=}1$ layer's recall denominator is the full GT person set, see Interpretability evidence), so $\Delta_{\mathrm{occ}}$ isolates the occlusion layer's contribution gap rather than a change in denominator. Per-layer mAP and gt\_frac are reported alongside.

\subsection{M2: Angle Error}

\textbf{Joint-angle definition.} Four joints are defined by COCO17 indices (Table~\ref{tab:joints}); the inner angle $\theta=\arccos\big((\mathbf{a}\cdot\mathbf{b})/(|\mathbf{a}|\,|\mathbf{b}|)\big)$, where $\mathbf{a}$ and $\mathbf{b}$ are the vectors from the vertex to the proximal and distal endpoints. Instances with $|\mathbf{a}|$ or $|\mathbf{b}|<10^{-6}$ are dropped.

\begin{table}[t]
\centering
\caption{Joint-angle definitions (COCO17 indices, 0-based).}
\label{tab:joints}
\renewcommand{\arraystretch}{1.15}
\begin{tabular}{lcc}
\toprule
\textbf{Joint} & \textbf{Vertex} & \textbf{Endpoints (proximal, distal)} \\
\midrule
left elbow & 7 & 5, 9 \\
right elbow & 8 & 6, 10 \\
left knee & 13 & 11, 15 \\
right knee & 14 & 12, 16 \\
\bottomrule
\end{tabular}
\end{table}

\textbf{Validity mask.} GT angle: all three GT points must have $v>0$. Predicted angle: the same three GT points must have $v>0$ and the predicted confidence must be $>0$---this keeps GT and predictions comparable on the same domain and avoids biasing toward missing predictions. Per-instance error $e=\theta_{\mathrm{pred}}-\theta_{\mathrm{gt}}$; we report RMSE, MAE, and P95, stratified by the vertex's GT visibility into $v{=}2$/$v{=}1$ layers, in linkage with M1.

\textbf{Aggregation and conditioning.} RMSE pools squared errors globally: $\mathrm{RMSE}=\sqrt{\frac{1}{n}\sum_i e_i^2}$ over all valid instances (left and right sides pooled at the instance level, never averaged per image or per side first). The arccos argument is clipped to $[-1,1]$ before the angle is computed, guarding against floating-point drift near $0^{\circ}$/$180^{\circ}$. Because the two models' valid-instance sets can differ slightly, we recomputed all 24 (dataset, joint, layer) RMSE comparisons on the intersection of the two sets as a robustness check: all 24 significance calls are unchanged (full numbers in the repository), so the comparison is not an artifact of differing coverage.

\textbf{Knee-valgus detection consistency.} The knee-valgus angle $\phi$ is defined as the image-plane angle between the hip$\rightarrow$knee vector and the hip$\rightarrow$ankle vector (how far the knee point deviates from the hip--ankle axis), computed only for instances where all three points have $v>0$. $\phi_{\mathrm{pred}}$ and $\phi_{\mathrm{GT}}$ are binarized at a threshold $\tau_\varphi$ (default $\tau_\varphi=15^{\circ}$, sensitivity analysis $\tau_\varphi\in\{10^{\circ},17^{\circ}\}$), and we report detection sensitivity, specificity, and Cohen's $\kappa$ with GT as the gold standard. Two clarifications: we do not report absolute detection rates (static datasets mix viewpoints, making absolute rates meaningless; consistency is what the protocol can measure); and $\phi$ is a 2D image-plane proxy under arbitrary viewpoints, whereas the FPPA literature values come from a standardized frontal camera~\cite{11}---our datasets mix viewpoints, so $\tau_\varphi$ takes only a magnitude reference from the literature and carries no clinical-diagnostic meaning.

\subsection{M3: Temporal Jitter}

\textbf{Data and matching.} PoseTrack18 val videos are inferred frame by frame independently (no tracking, no cross-frame information); per-frame predictions are matched to GT by Hungarian matching on full-keypoint OKS (same convention as M1). Segments break when the frame gap exceeds 1; only track segments with $\ge$10 consecutive matched frames are retained. This retention rule is strict: 94--95\% of raw segments are shorter than 10 frames and are dropped, and the retained segments cover 53.1--53.4\% of annotated frames (round5-3: 704 segments over 4,767 frames; YOLO11s-pose: 728 over 4,739); in addition, 16.3--20.1\% of GT person instances are never matched to any prediction. M3 numbers therefore describe the matched, retained subpopulation rather than the whole video. All displacements are normalized by the GT person-box scale $s_t=\sqrt{w\cdot h}$ (consistent with the OKS scale convention).

\textbf{Three metrics} (computed per keypoint and per track segment, then averaged):

\begin{enumerate}
\item frame-to-frame displacement $d_t=\lVert\mathbf{p}_t-\mathbf{p}_{t-1}\rVert_2/s_t$: report mean and P95;
\item jitter $j_t=\lVert\mathbf{p}_t-\hat{\mathbf{s}}_t\rVert_2/s_t$, where $\hat{\mathbf{s}}$ is the trajectory after a causal 5-frame median filter: targets high-frequency noise (core metric);
\item jump rate: fraction of frames with $d_t>\tau_j$, default $\tau_j=0.05$ (5\% of person-box scale), sensitivity $\tau_j\in\{0.10,0.20\}$.
\end{enumerate}

\textbf{Before/after OneEuro comparison.} A causal OneEuro filter is applied offline to $\mathbf{p}_t$ with parameters fixed at Casiez's defaults (min\_cutoff$=1.0$, beta$=0.007$, d\_cutoff$=1.0$)~\cite{13}. The three metrics and the matched-scope AP are recomputed after smoothing, and the accuracy cost is reported. Two filtering domains are used: the normalized-coordinate domain (q domain, the protocol's original design) and the pixel domain. The empirical finding (consistent across both models): filtering in the q domain re-injects the GT person box's per-frame scale noise ($|s_t/s_{t-1}-1|$, median 0.97\%, P95 8.1\%) into the pixel domain during de-normalization, so jitter metrics rise rather than fall and AP drops slightly; direct pixel-domain filtering costs zero AP and lowers all three metrics. The default-parameter conclusion is insensitive to tuning: on a 12-point grid (min\_cutoff $\in\{0.5,1,2,4\}\times$ beta $\in\{0.001,0.007,0.05\}$), every pixel-domain setting lowers all seven M3 metrics below the raw baseline for both models, with $|\Delta\mathrm{AP}|\le 0.0004$; the q domain at the strongest smoothing loses up to 0.0071 AP (grid in the repository). \textbf{We adopt the pixel domain as the primary convention}, and report the q domain as a sensitivity convention (Table~\ref{tab:m3}, two columns).

\subsection{Statistical Conventions}

\begin{itemize}
\item Model comparison: M1/M2 resample at image level and M3 at track-segment level, 10,000 bootstrap replicates (seed 42); we report 95\% percentile CIs of the difference $\Delta$ and judge significance by whether the CI excludes 0. M1 is a paired design (shared image universe $=$ the union of GT and predicted images of both models); the valid unit sets of M2/M3 differ with matching results, so they use independent resampling, as noted in the protocol.
\item The comparison list is fixed before the experiments (2 models $\times$ 3 datasets $\times$ layers/joints/metrics, plus the RQ5 oracle list of 2 transformer models $\times$ 3 datasets $\times$ layers/joints); every entry is reported with its CI, with no selection of favorable results.
\item RQ4 (protocol economy): on (dataset $\times$ layer) cells we compute the sign-consistency $\tau_s=(C-D)/(C+D)$ ($C$ concordant pairs, $D$ discordant pairs) between protocol-metric differences and full-mAP differences, and the Spearman $\rho$ between the two difference vectors. $\tau_s$ counts concordant pairs among all pairwise cell comparisons, excluding pairs whose two cells share an identical full-mAP difference (a tie, for which sign consistency is undefined): at $n=6$ (3 datasets $\times$ 2 layers), 15 pairwise comparisons minus the 3 same-dataset layer pairs $=$ 12 reported pairs; at the joint level ($n=24$: 3 datasets $\times$ 4 joints $\times$ 2 layers), 276 pairwise comparisons minus 84 same-dataset same-joint layer pairs (3 $\times$ 28) $=$ 192 reported pairs. Both analyses are exploratory, and the sample size is stated in the main text. The $n=6$ cells contain only three independent dataset-level values per metric (the two models share the same three datasets), so the effective unit is the dataset and the correlation analysis is exploratory rather than confirmatory. All intervals are per-comparison 95\% CIs without multiple-comparison correction; significance serves only as a secondary aid, and the core conclusions rest on three-dataset directional consistency and effect sizes (a Benjamini--Hochberg FDR sensitivity analysis over the core comparison families is in Appendix~A, Table~A14).
\end{itemize}

\subsection{Harness and Reproducibility}

The execution layer is an open-source eval harness: \texttt{model.val(save\_json=True)} produces official metrics and prediction JSONs in a single inference pass; the M1/M2/M3 engines are purely offline (NumPy) and do not touch the GPU. Three gates guarantee the conventions are implemented correctly (see 3.2). The reproducibility contract (environment, weights, data, step-by-step commands, convention notes) is in the repository's REPRODUCIBILITY.md; all model weights (including round5-3 and the YOLO-oracle variants) are distributed via the repository's GitHub Releases, with checksums and acquisition instructions recorded in REPRODUCIBILITY.md. The protocol's JSON contract (schema\_version$=$1) is available for downstream plotting and secondary analysis. The harness and this paper's source are open-sourced at \url{https://github.com/fanyifan090409-dev/pose-eval-harness} and \url{https://github.com/fanyifan090409-dev/elthan-pose-research}; the training fork (ultralytics-based) is at \url{https://github.com/fanyifan090409-dev/elthan-pose}.

\section{Experimental Setup}

\textbf{Datasets.} Four public datasets: COCO val2017 (general-domain control; $v$ distribution 55.4\%/7.8\%/36.8\% for $v{=}2/1/0$), CrowdPose val (crowded/occlusion domain; $v{=}1$ marks real occlusion; distribution 22.5\%/30.8\%/remainder), OCHuman test (heavy-occlusion domain; distribution 69.2\%/14.3\%/remainder), PoseTrack18 val (video temporal; 170 videos / 8,923 annotated frames; GT $v\in\{0,1\}$, where $v{=}0$ is a (0,0) placeholder). All datasets are unified under the COCO17 keypoint format. PoseTrack was obtained under MPII's official academic license and is used for academic evaluation only.

\textbf{Models.} Two baseline models: (1) the official YOLO11s-pose pretrained weights~\cite{14}; (2) round5-3, obtained by fine-tuning YOLO11s-pose on public datasets including COCO, MPII, CrowdPose, and OCHuman, representing training with occlusion-domain data (the name is the author's internal run identifier: the third checkpoint of the fifth data-configuration round of the training pipeline). The two share the same architecture and inference configuration and differ only in training data, making them suitable for testing whether the protocol is sensitive to training-data differences. round5-3's training data is restricted to training splits (COCO train2017, MPII train, CrowdPose train, OCHuman train, plus OpenThermalPose train at controlled proportion~\cite{18}); none of the four evaluation splits used in this paper (COCO val2017, CrowdPose val, OCHuman test, PoseTrack18 val) appears in its training data, so the protocol numbers are leak-free. Two further models extend the baseline across architectures: (3) ViTPose-B and (4) ViTPose-L~\cite{17}, top-down transformer models of the same architecture family with different capacity (ViT-B/L backbones), evaluated under a detector-free GT-bbox oracle protocol. To make every cross-family comparison a same-protocol test, the two YOLO models are additionally evaluated under the oracle protocol: each GT person box is cropped, letterboxed to $640\times640$, and passed through the single-stage model without flip test. Because this preprocessing differs from the transformer family's UDP $256\times192$ input, the YOLO-oracle numbers are not directly comparable to their detection-based numbers, but they are fully comparable to the ViTPose oracle numbers. The four models thus span two architecture families and support three comparisons: training-data effects within the YOLO family (RQ1--RQ4), architecture-capacity effects within the ViTPose family (RQ5), and cross-family effects under the shared oracle protocol (RQ5).

\textbf{Inference and conventions.} $640\times640$ input, frame-by-frame independent inference, predictions saved with save\_json. All mAP numbers use the official released-numbers convention of pycocotools 2.0.11, whose doubled-bbox fallback scores a GT keypoint with $vg>0$ but zero keypoint area ($k_1=0$, the COCO (0,0) placeholder) using a bbox-derived scale rather than dropping it. Note that this convention can differ from ultralytics' internal evaluator by more than 20 points (sources: GTs with num\_keypoints$=$0 marked ignore, the OKS doubled-bbox fallback, maxDets$=$20); every table in the paper states its convention. PoseTrack has no official pycocotools reference, so its official numbers are reported only under the internal-evaluator convention (round5-3 0.1085 / YOLO11s-pose 0.1213 mAP$^{50-95}$(P)). Gate 1 (agreement with the official library) measured $\le 0.0001$ on all six model--dataset cells (exactly 0 on five).

\textbf{Detector-free oracle protocol.} For the ViTPose models, each GT person box is used directly as the detection result---one prediction per GT annotation---so that no detector error enters the pipeline; preprocessing follows the official top-down convention (UDP affine transform, $256\times192$ input, flip test); PoseTrack's 657 degenerate $0\times0$ GT boxes are skipped at source. The oracle protocol isolates the pose head: all remaining error is attributable to keypoint localization and visibility reasoning, not to detection. Gate 1 (agreement with pycocotools 2.0.11) measures exactly 0 on all six ViTPose model--dataset cells, and ViTPose-B's full COCO AP (0.7676) is 0.96 points higher than the official paper number (75.8, detector-box input convention)---direction and magnitude as expected, because the oracle replaces the detector box with the GT person box and removes detector error~\cite{17}. Under this protocol all four models are evaluated on equal footing; the YOLO-oracle variant differs only in preprocessing (GT-bbox crop plus letterbox instead of UDP plus flip), which Section~5.5 reports and discusses.

\section{Results}

\subsection{RQ1: Full mAP Masks Occlusion Differences}

\begin{figure*}[t]
\centering
\begin{minipage}[b]{0.32\linewidth}
  \centering
  \includegraphics[width=\linewidth]{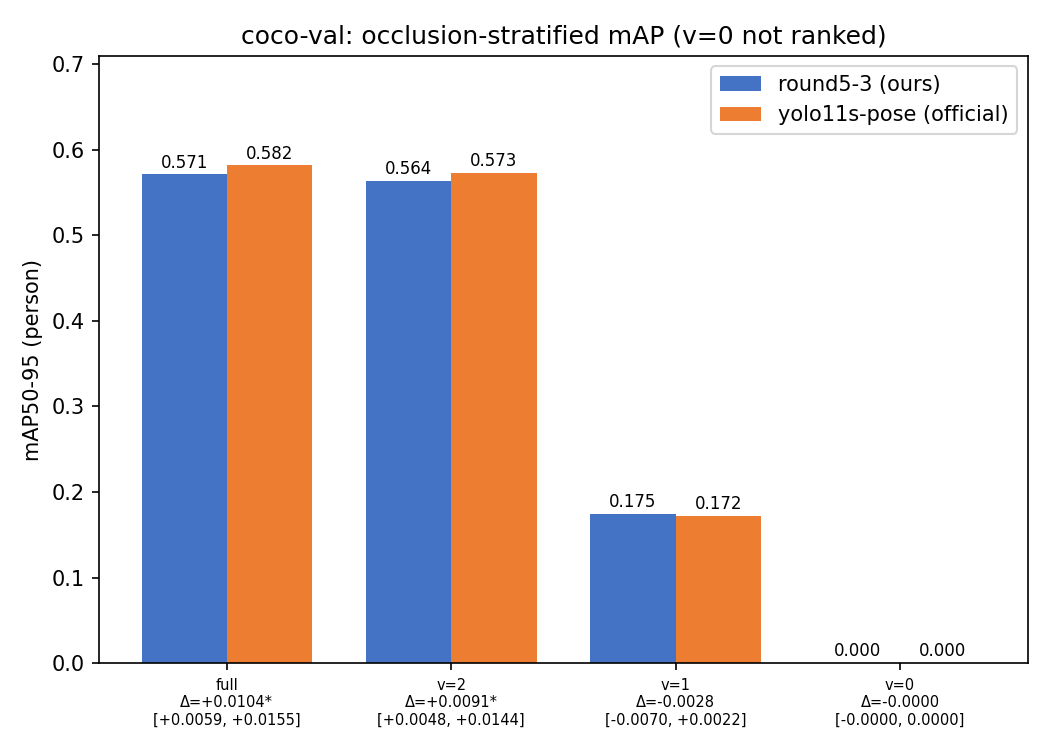}
\end{minipage}\hfill
\begin{minipage}[b]{0.32\linewidth}
  \centering
  \includegraphics[width=\linewidth]{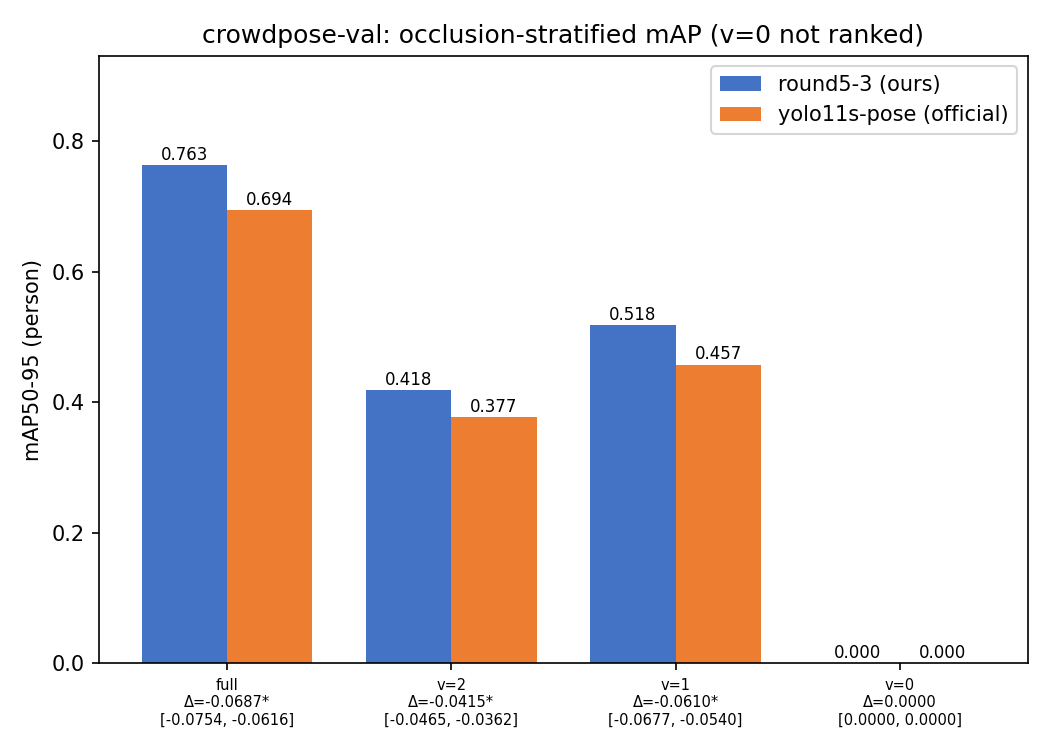}
\end{minipage}\hfill
\begin{minipage}[b]{0.32\linewidth}
  \centering
  \includegraphics[width=\linewidth]{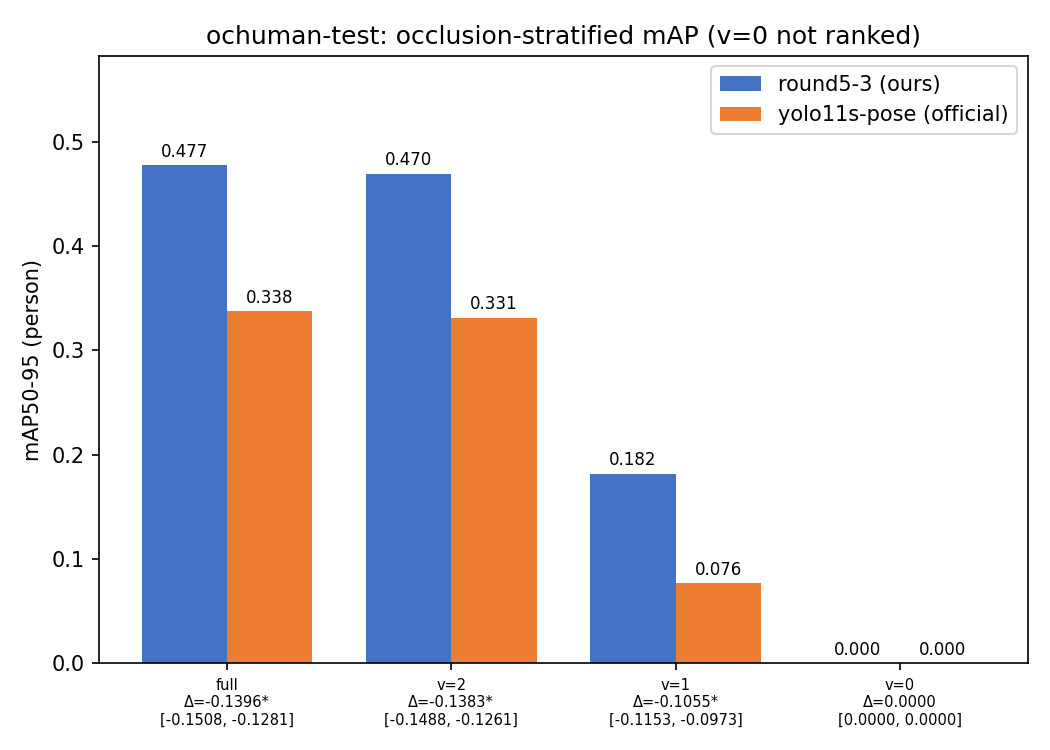}
\end{minipage}
\caption{M1 occlusion-stratified mAP$^{50-95}$(P) per dataset: (a) COCO val2017, (b) CrowdPose val, (c) OCHuman test. Bars show full and per-layer mAP for round5-3 and YOLO11s-pose; annotations below each pair give the model difference $\Delta$ with its 95\% bootstrap CI ($*$ $=$ CI excludes 0). The $v{=}0$ layer is shown for completeness and excluded from ranking.}
\label{fig:m1}
\end{figure*}

Table~\ref{tab:m1} gives the stratified mAP of the two models on three datasets, together with the layer-representativeness evidence gt\_frac and gt\_person\_frac; Fig.~\ref{fig:m1} visualizes the stratification. Three observations.

\begin{table}[t]
\centering
\caption{M1 stratified mAP$^{50-95}$(P) (pycocotools 2.0.11 convention). Bold $=$ significantly better (bootstrap CI excludes 0). gt\_frac $=$ fraction of GT keypoints in the layer; gt\_person\_frac $=$ fraction of GT persons possessing at least one keypoint of the layer (both are GT properties, identical for the two models). $\Delta_{\mathrm{occ}}=$ mAP($v{=}1$ layer) $-$ mAP(full) on the same full GT person set (Section 3.2); its bootstrap CIs are in Appendix~A. --- $=$ not defined for this layer. CrowdPose's official annotations mark nearly all visible points as $v{=}2$; together with the full-person recall denominator (gt\_person\_frac 0.591) this systematically underestimates its v2-layer AP; cross-layer comparisons use $\Delta_{\mathrm{occ}}$ (the denominator bias cancels in the difference).}
\label{tab:m1}
\renewcommand{\arraystretch}{1.15}
\resizebox{\columnwidth}{!}{%
\begin{tabular}{lccccccc}
\toprule
\textbf{Dataset} & \textbf{Layer} & \textbf{round5-3} & \textbf{YOLO11s-pose} & \textbf{$\Delta_{\mathrm{occ}}$ (r5-3)} & \textbf{$\Delta_{\mathrm{occ}}$ (YOLO)} & \textbf{gt\_frac} & \textbf{gt\_person\_frac} \\
\midrule
COCO val & full & 0.5712 & \textbf{0.5816} & --- & --- & 1.000 & 1.000 \\
COCO val & $v{=}2$ & 0.5637 & \textbf{0.5728} & --- & --- & 0.554 & 0.988 \\
COCO val & $v{=}1$ & 0.1747 & 0.1719 & $-0.3965$ & $-0.4097$ & 0.078 & 0.564 \\
COCO val & $v{=}0$ & 0 & 0 & --- & --- & 0.368 & 0.947 \\
CrowdPose val & full & \textbf{0.7626} & 0.6938 & --- & --- & 1.000 & 1.000 \\
CrowdPose val & $v{=}2$ & \textbf{0.4183} & 0.3768 & --- & --- & 0.225 & 0.591 \\
CrowdPose val & $v{=}1$ & \textbf{0.5183} & 0.4573 & $-0.2443$ & $-0.2366$ & 0.308 & 0.776 \\
CrowdPose val & $v{=}0$ & 0 & 0 & --- & --- & 0.467 & 1.000 \\
OCHuman test & full & \textbf{0.4775} & 0.3379 & --- & --- & 1.000 & 1.000 \\
OCHuman test & $v{=}2$ & \textbf{0.4696} & 0.3314 & --- & --- & 0.692 & 1.000 \\
OCHuman test & $v{=}1$ & \textbf{0.1819} & 0.0765 & $-0.2955$ & $-0.2614$ & 0.143 & 0.736 \\
OCHuman test & $v{=}0$ & 0 & 0 & --- & --- & 0.164 & 0.902 \\
\bottomrule
\end{tabular}}
\end{table}

First, \textbf{$\Delta_{\mathrm{occ}}$ is negative everywhere and large in magnitude}: $-0.40$ and $-0.41$ on COCO, $-0.24$ on CrowdPose, and $-0.26$ to $-0.30$ on OCHuman. In every combination tested, full mAP overestimates occluded-scene performance by 24--41 points. Part of this gap is inherent: $v{=}1$ keypoints are harder for any model, so $\Delta_{\mathrm{occ}}$ conflates inherent difficulty with masking. The model-level quantity that isolates masking is $\Delta\Delta_{\mathrm{occ}}=\Delta_{\mathrm{occ}}(\mathrm{YOLO})-\Delta_{\mathrm{occ}}(\mathrm{r5\text{-}3})$, which is significant on all three datasets (COCO $-0.013$ [$-0.018$, $-0.009$], CrowdPose $+0.008$ [$+0.004$, $+0.011$], OCHuman $+0.034$ [$+0.024$, $+0.043$]; Appendix~A, Table~A2): the two models diverge in how much their full mAP overstates occluded-scene accuracy, and that divergence is invisible in the aggregate score.

Second, \textbf{stratification changes the strength of model-comparison conclusions}. On general-domain COCO, YOLO11s-pose is significantly better in full and $v{=}2$-layer mAP ($\Delta=+0.0104$, 95\% CI $[0.0059,0.0155]$), but we found no evidence of a $v{=}1$-layer difference ($\Delta=-0.0028$, ns)---the significant advantage in the aggregate comes almost entirely from visible keypoints. On the occlusion-domain datasets, round5-3 is significantly better in all three layers (CrowdPose full $\Delta=-0.0687$, OCHuman full $\Delta=-0.1396$, 95\% CIs exclude 0), and the gain from occlusion-domain training is equally significant in the $v{=}1$ layer (OCHuman v1 $\Delta=-0.1055$; CrowdPose's v2-layer AP is underestimated by the full-person recall denominator, and its cross-layer conclusions rest on $\Delta_{\mathrm{occ}}$, Section~3.2). Sign-consistency with full mAP is nevertheless perfect ($\tau_s=1.0$, Section~5.4): for the two models tested, stratification changes the magnitude of the model gap, not its direction. We extend the same stratification to two transformer models and to the YOLO pair itself under the oracle protocol (Section~5.5) and find the same within-family stability; cross-family re-ranking under the common oracle protocol is tested in Section~5.5.

Third, \textbf{the $v{=}0$ layer is uniformly 0} (both models, three datasets), confirming the protocol's expectation: missing keypoints cannot be estimated; the layer is kept as completeness evidence, not for ranking.

\subsection{RQ2: Angle Error and Knee-Valgus Detection Consistency}

\begin{figure*}[t]
\centering
\begin{minipage}[b]{0.32\linewidth}
  \centering
  \includegraphics[width=\linewidth]{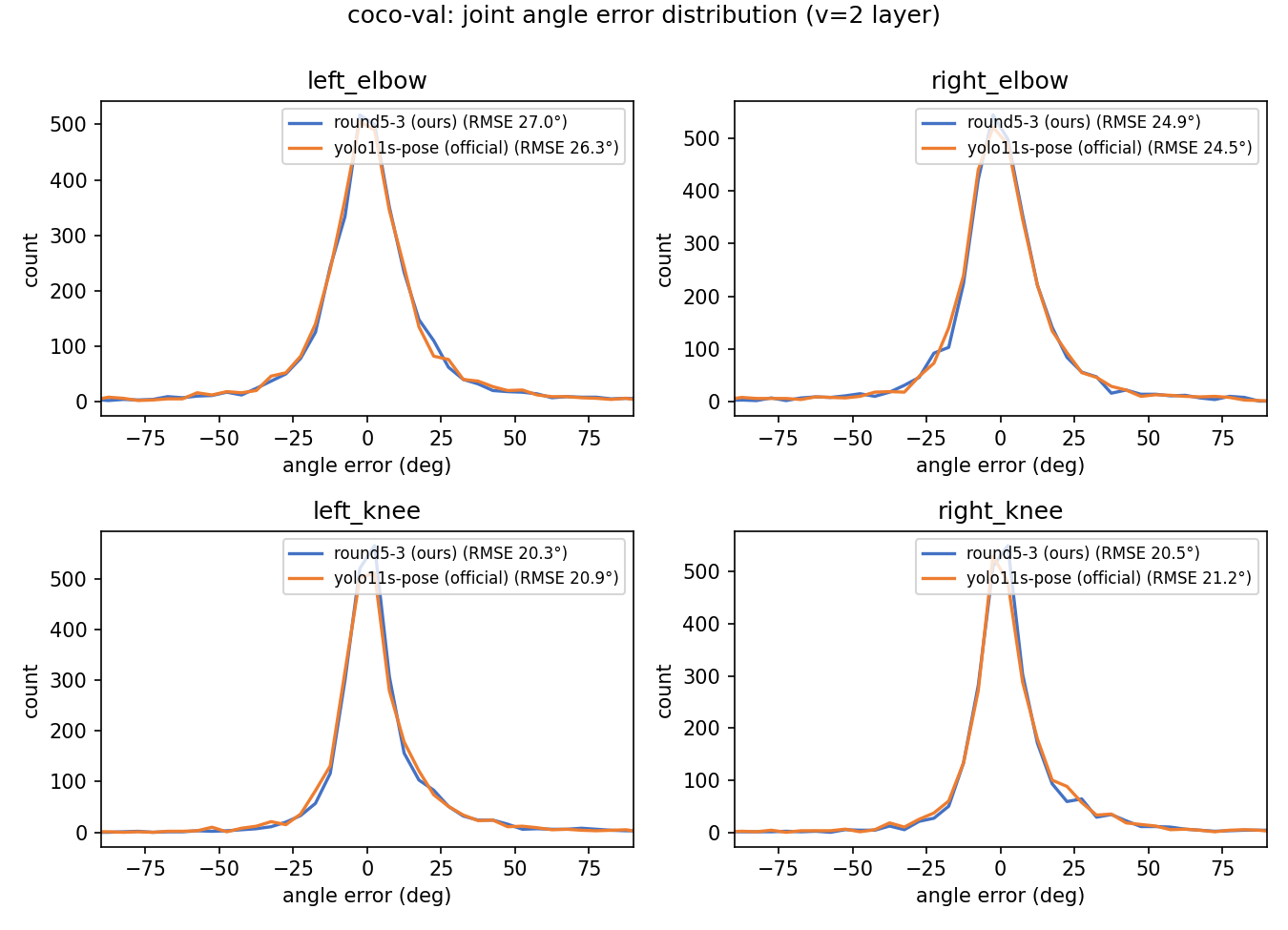}
\end{minipage}\hfill
\begin{minipage}[b]{0.32\linewidth}
  \centering
  \includegraphics[width=\linewidth]{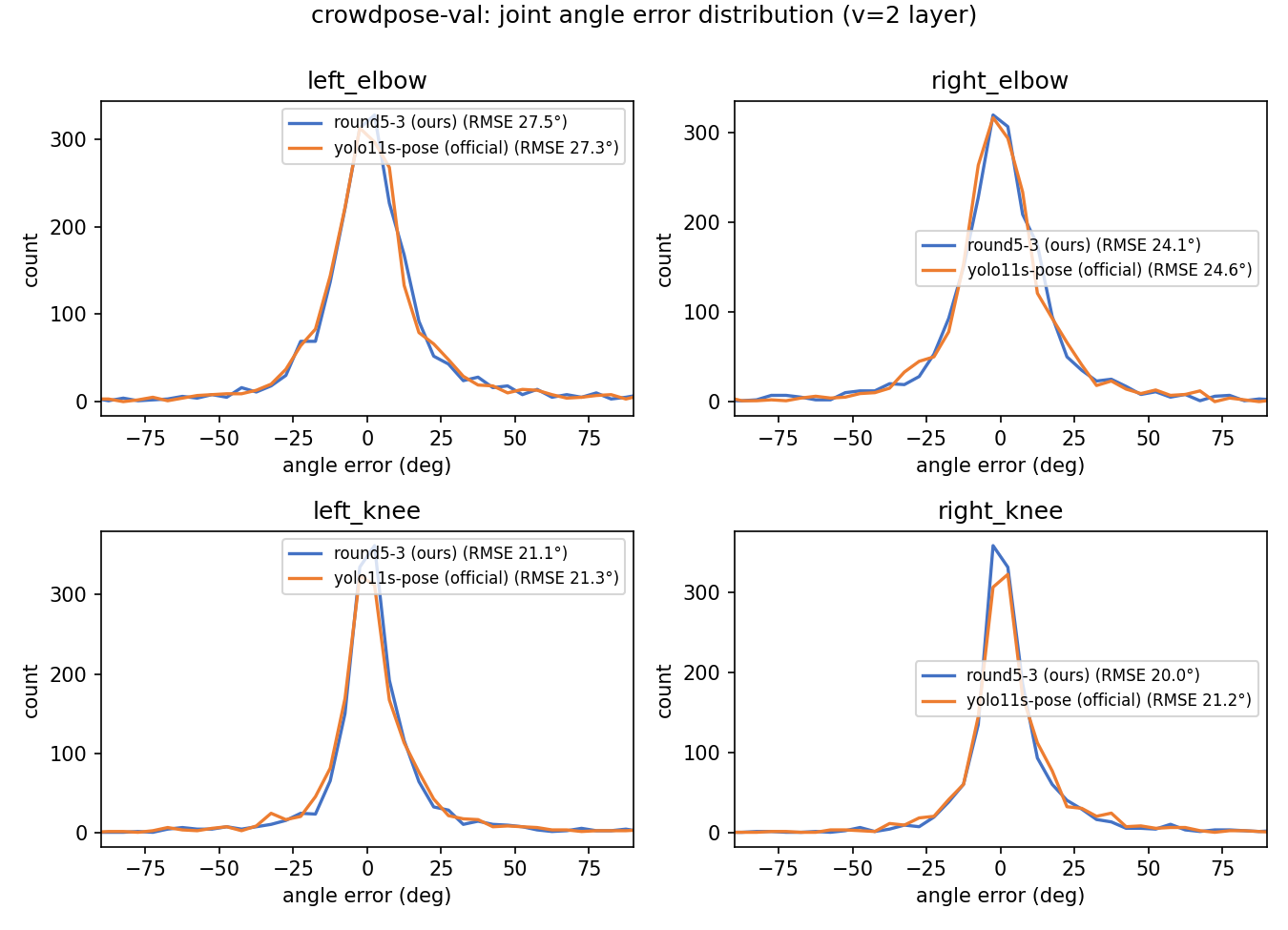}
\end{minipage}\hfill
\begin{minipage}[b]{0.32\linewidth}
  \centering
  \includegraphics[width=\linewidth]{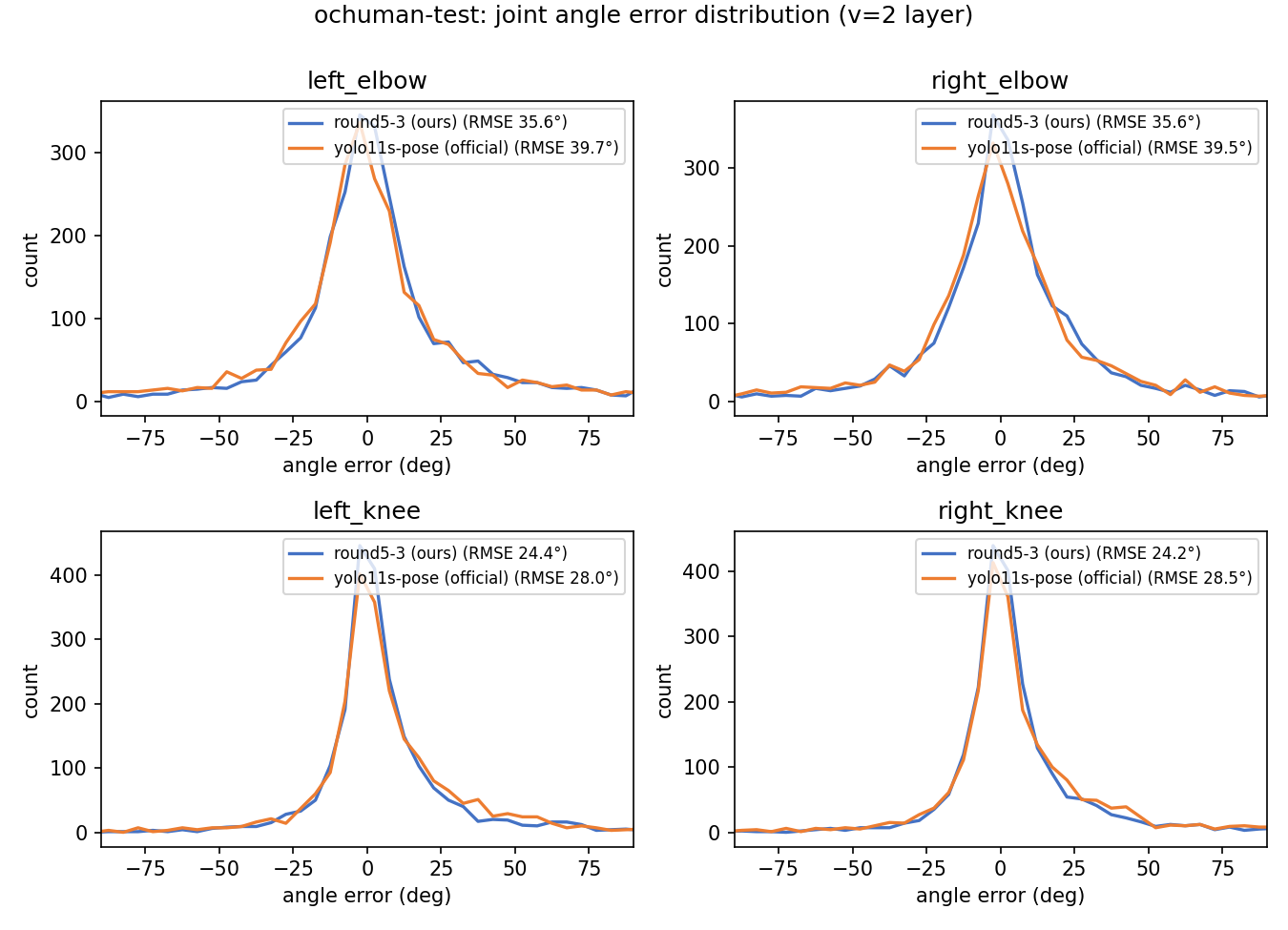}
\end{minipage}
\caption{M2 joint-angle error distributions ($v{=}2$ layer) per dataset: (a) COCO val2017, (b) CrowdPose val, (c) OCHuman test. Each panel shows per-joint error histograms ($\theta_{\mathrm{pred}}-\theta_{\mathrm{gt}}$) for both models, with RMSE in the legend. Distributions concentrate near $0^{\circ}$; the occlusion-domain datasets separate the two models more clearly (Table~\ref{tab:m2}).}
\label{fig:m2}
\end{figure*}

Table~\ref{tab:m2} gives elbow/knee angle RMSE with left and right sides pooled at the instance level (stratified by $v{=}2$/$v{=}1$); Fig.~\ref{fig:m2} shows the per-joint error distributions ($v{=}2$ layer). Three observations.

\begin{table}[t]
\centering
\caption{M2 joint-angle RMSE ($^{\circ}$, instances pooled per side). Bold $=$ significantly better (bootstrap CI excludes 0); per-comparison CIs are in Appendix~A (Table~A3).}
\label{tab:m2}
\renewcommand{\arraystretch}{1.15}
\resizebox{\columnwidth}{!}{%
\begin{tabular}{lccccc}
\toprule
\textbf{Dataset} & \textbf{Model} & \textbf{Elbow $v{=}2$} & \textbf{Elbow $v{=}1$} & \textbf{Knee $v{=}2$} & \textbf{Knee $v{=}1$} \\
\midrule
COCO val & round5-3 & 26.0 & 35.9 & 20.4 & 29.2 \\
COCO val & YOLO11s-pose & 25.4 & 35.6 & 21.0 & 31.2 \\
CrowdPose val & round5-3 & 25.9 & \textbf{31.6} & 20.6 & \textbf{25.9} \\
CrowdPose val & YOLO11s-pose & 26.0 & 35.0 & 21.3 & 30.5 \\
OCHuman test & round5-3 & \textbf{35.6} & \textbf{45.7} & \textbf{24.3} & \textbf{32.0} \\
OCHuman test & YOLO11s-pose & 39.6 & 57.7 & 28.3 & 39.7 \\
\bottomrule
\end{tabular}}
\end{table}

First, \textbf{angle error is systematically amplified by occlusion}: in every model--dataset combination, the $v{=}1$-layer RMSE point estimates exceed the $v{=}2$-layer RMSE (elbow $+5.7^{\circ}$ to $+18.1^{\circ}$, knee $+5.3^{\circ}$ to $+11.4^{\circ}$), with the largest layer gaps in the heavy-occlusion domain (OCHuman); per-combination paired bootstrap CIs of the v1$-$v2 layer differences are in Appendix~A (Table~A13). The complementary MAE and P95 statistics are reported per model, dataset, joint, and layer in Appendix~A (Table~A15); they reproduce the same pattern (in every one of the 24 model--dataset--joint combinations, the $v{=}1$ cell exceeds its $v{=}2$ counterpart in both MAE and P95, with the largest errors in the OCHuman domain). This aligns with M1's $\Delta_{\mathrm{occ}}$ observation; the two corroborate each other.

Second, \textbf{the angle metrics separate information that full mAP cannot}. On general-domain COCO we found no evidence of an angle difference between the two models (consistent with the 1-point mAP gap); but on the occlusion domains round5-3 is more accurate in the $v{=}1$ layer in every tested comparison: CrowdPose elbow 31.6$^{\circ}$ vs.\ 35.0$^{\circ}$, knee 25.9$^{\circ}$ vs.\ 30.5$^{\circ}$ (both significant); OCHuman elbow 45.7$^{\circ}$ vs.\ 57.7$^{\circ}$, knee 32.0$^{\circ}$ vs.\ 39.7$^{\circ}$ (all significant). Note that on OCHuman the $v{=}2$-layer angle difference is also significant, indicating that the benefit of occlusion-domain training is not limited to occluded keypoints themselves. These comparisons use each model's own valid-instance set; recomputed on the intersection of the two sets, all 24 significance calls are unchanged (Section~3.3).

Third, \textbf{knee-valgus detection consistency} (Table~\ref{tab:kappa}): round5-3's $\kappa$ exceeds YOLO11s-pose's on all three datasets at the default threshold $\tau_\varphi{=}15^{\circ}$ (COCO 0.759 vs.\ 0.745; CrowdPose 0.739 vs.\ 0.669; OCHuman 0.602 vs.\ 0.519), and the conclusion is stable under threshold sensitivity analysis ($\tau_\varphi\in\{10^{\circ},17^{\circ}\}$). Bootstrap CIs (Appendix~A, Table~A4) establish the CrowdPose and OCHuman gaps as significant ($\Delta\kappa$ $-0.069$ [$-0.098$, $-0.041$] and $-0.084$ [$-0.121$, $-0.046$]), whereas the COCO gap is directionally consistent but not significant ($\Delta\kappa$ $-0.014$ [$-0.045$, $0.016$]). Both models' $\kappa$ on OCHuman are low ($<0.61$), suggesting that $\phi$'s measurement noise rises under heavy occlusion.

\begin{table}[t]
\centering
\caption{Knee-valgus detection consistency (binarized $\phi$ vs.\ GT). Bold $=$ higher value (round5-3); bootstrap CIs for $\kappa$ and for the model difference are in Appendix~A.}
\label{tab:kappa}
\renewcommand{\arraystretch}{1.15}
\resizebox{\columnwidth}{!}{%
\begin{tabular}{lccccc}
\toprule
\textbf{Dataset} & \textbf{Model} & \textbf{$\tau_\varphi{=}10^{\circ}$ $\kappa$} & \textbf{$\tau_\varphi{=}15^{\circ}$ $\kappa$} & \textbf{$\tau_\varphi{=}17^{\circ}$ $\kappa$} & \textbf{$\tau_\varphi{=}15^{\circ}$ sens./spec.} \\
\midrule
COCO val & round5-3 & 0.724 & \textbf{0.759} & 0.773 & 0.790/0.950 \\
COCO val & YOLO11s-pose & 0.697 & 0.745 & 0.759 & 0.783/0.943 \\
CrowdPose val & round5-3 & 0.711 & \textbf{0.739} & 0.731 & 0.785/0.937 \\
CrowdPose val & YOLO11s-pose & 0.666 & 0.669 & 0.665 & 0.722/0.926 \\
OCHuman test & round5-3 & 0.601 & \textbf{0.602} & 0.603 & 0.646/0.926 \\
OCHuman test & YOLO11s-pose & 0.529 & 0.519 & 0.502 & 0.592/0.900 \\
\bottomrule
\end{tabular}}
\end{table}

\subsection{RQ3: Temporal Stability of Tracking-Free Inference and the Cost of Smoothing}

\begin{figure}[t]
\centering
\includegraphics[width=\columnwidth]{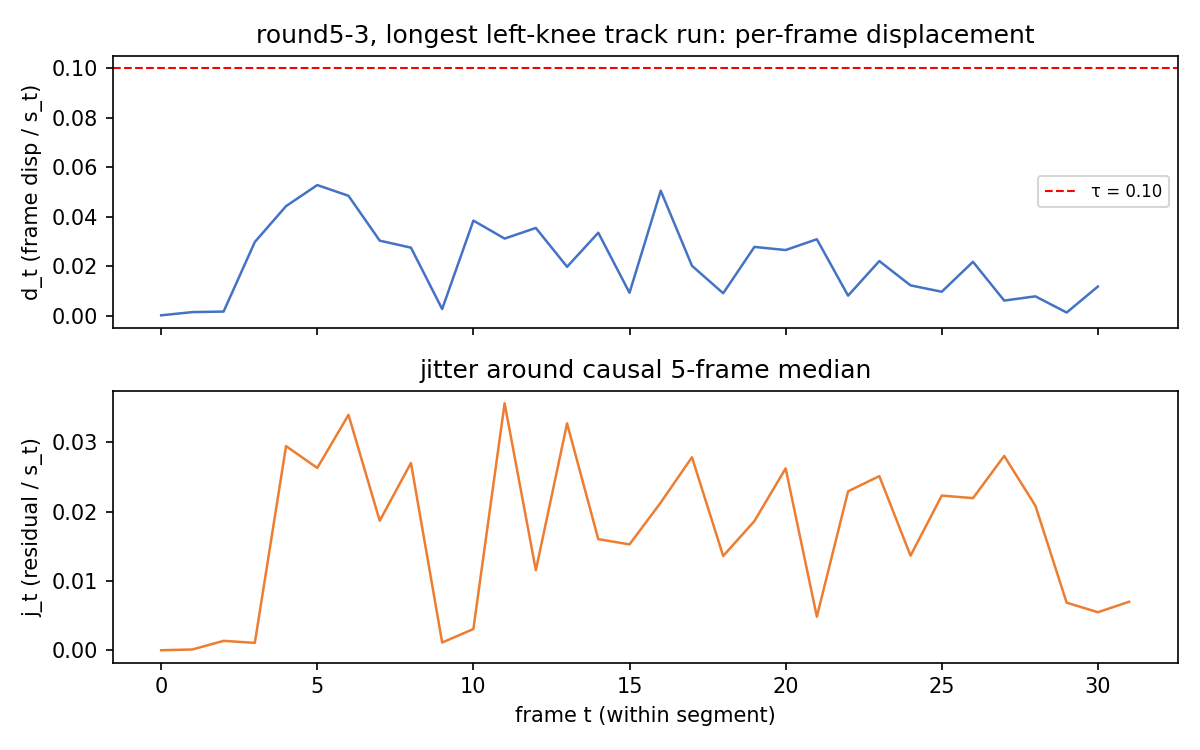}
\caption{M3 temporal jitter on an example PoseTrack18 val track (round5-3's longest left-knee retained segment; for illustration, not a summary statistic): (top) per-frame displacement $d_t$ normalized by person-box scale, with the sensitivity jump threshold $\tau_j=0.10$; (bottom) residual jitter $j_t$ around a causal 5-frame median filter.}
\label{fig:m3}
\end{figure}

On PoseTrack18 val the two models yield 704 / 728 track segments (161 / 160 videos, median segment length 31 frames). The segment-retention filter is strict: these segments cover 53.4\% / 53.1\% of annotated frames, and 20.1\% / 16.3\% of GT person instances are never matched (Section~3.4). Table~\ref{tab:m3} gives the overall three metrics after person-scale normalization, plus the AP cost of smoothing; Fig.~\ref{fig:m3} shows an example track (round5-3's longest left-knee retained segment, chosen as a visualization example; the summary statistics are over all retained segments).

\begin{table}[t]
\centering
\caption{M3 temporal metrics (overall, averaged over 17 keypoints) and smoothing AP cost. Matched-scope AP is the AP recomputed over the same matched track frames used for the jitter metrics (raw values: round5-3 0.1316 / YOLO11s-pose 0.1408); --- $=$ raw reference, no smoothing applied. Bold AP cost $=$ within $\pm0.0002$ of zero (zero accuracy cost). Per-metric domain-difference and between-model CIs are in Appendix~A (Table~A5).}
\label{tab:m3}
\renewcommand{\arraystretch}{1.15}
\resizebox{\columnwidth}{!}{%
\begin{tabular}{lcccccc}
\toprule
\textbf{Model} & \textbf{Convention} & \textbf{disp\_mean} & \textbf{jitter\_mean} & \textbf{jump\_0.05} & \textbf{jump\_0.10} & \textbf{AP cost} \\
\midrule
round5-3 & raw & 0.0409 & 0.0521 & 0.2062 & 0.0726 & --- \\
round5-3 & pixel-domain smoothing & 0.0380 & 0.0509 & 0.1870 & 0.0634 & \textbf{$+0.0001$} \\
round5-3 & q-domain smoothing & 0.0432 & 0.0528 & 0.2200 & 0.0807 & $-0.0013$ \\
YOLO11s-pose & raw & 0.0430 & 0.0535 & 0.2153 & 0.0777 & --- \\
YOLO11s-pose & pixel-domain smoothing & 0.0399 & 0.0523 & 0.1943 & 0.0679 & \textbf{$+0.0002$} \\
YOLO11s-pose & q-domain smoothing & 0.0462 & 0.0546 & 0.2336 & 0.0878 & $-0.0029$ \\
\bottomrule
\end{tabular}}
\end{table}

Three observations. First, \textbf{we found no evidence of a difference in raw temporal stability between the two models}: the bootstrap CIs of all seven metrics (disp/jitter means and P95s, three jump-rate thresholds) include 0. We found no evidence that occlusion-domain training data changed temporal stability---``no evidence that occlusion training affects temporal behavior'' is itself an informative result, suggesting that data-side interventions should be considered separately from temporal post-processing.

Second, \textbf{pixel-domain OneEuro smoothing buys stability at zero matched-scope AP cost}: the jump rate ($\tau_j=0.05$) drops by 9.3\% (round5-3) and 9.7\% (YOLO11s-pose) relative---the more lenient sensitivity threshold $\tau_j=0.10$ drops by even more (12.7\% / 12.6\%, Table~\ref{tab:m3}), so the benefit is not an artifact of the default threshold---frame-to-frame displacement mean drops by about 7\%, and AP changes stay within $\pm0.0002$. The benefit is not an artifact of the default filter parameters: all 12 settings of a OneEuro grid (min\_cutoff $\times$ beta, Section~3.4) lower every M3 metric below the raw baseline for both models, at $|\Delta\mathrm{AP}|\le 0.0004$. Third, \textbf{q-domain smoothing backfires}: all three metrics rise (jump\_0.05 to 0.2200/0.2336) and AP drops slightly---the GT person box's per-frame scale noise is re-injected into the pixel domain during de-normalization (mechanism in 3.4). This comparison shows that when smoothing tracking-free per-frame outputs, the choice of filtering domain decides the benefit; the protocol therefore fixes the pixel domain as the primary convention.

\subsection{RQ4: Correlations Among Metrics}

On the (dataset $\times$ layer) cells ($n=6$), we compute the sign-consistency $\tau_s$ between protocol-metric differences ($\Delta=$ YOLO11s-pose $-$ round5-3, with RMSE differences sign-flipped so that positive always means YOLO11s-pose is better) and full-mAP differences, counting only the 12 pairs with distinct full-mAP differences (Section~3.5): stratified mAP $\tau_s=1.0$ (12/12 pairs), elbow-angle RMSE $\tau_s=1.0$ (12/12), knee-angle RMSE $\tau_s=0.667$ (10/12); extended to 24 joint-level cells, $\tau_s=0.719$ ($C{=}165$, $D{=}27$, 192 pairs total). Spearman $\rho$ between the same difference vectors: full $\Delta$ vs.\ stratified $\Delta$, $\rho=0.956$; full $\Delta$ vs.\ elbow angle, $\rho=0.956$; full $\Delta$ vs.\ knee angle, $\rho=0.717$; elbow vs.\ knee, $\rho=0.771$ ($n=6$ cells over only 3 independent datasets; the p-values are nominal only, $p=0.003/0.003/0.109/0.072$, and are not used for inference). The two identical $\rho=0.956$ values are not a transcription error: the stratified-mAP and elbow-angle difference vectors rank the six cells identically, and the only rank differences between either of them and the full-mAP differences occur within the two layers of the same dataset, which share an identical full-mAP difference (a tie), leaving Spearman's numerator unchanged in both cases.

Two interpretations (exploratory; $n=6$ cells over only 3 independent datasets). First, \textbf{stratified mAP's ranking information is perfectly sign-consistent with full mAP ($\tau_s=1.0$)}; its protocol value lies not in re-ranking but in magnitude diagnosis---the ``how much is masked'' that $\Delta_{\mathrm{occ}}$ provides is something a scalar mAP cannot give. Second, \textbf{the angle metrics carry partially independent information}: the knee-angle RMSE and full-AP difference vectors do not rank the cells identically, and on COCO the directions diverge (round5-3 has better knee angles yet lower full mAP; Table~\ref{tab:m2})---the angle protocol captures a dimension of difference beyond OKS (descriptive evidence, not a statistical inference). For protocol economy: stratified mAP and angle error are not redundant; each captures a different blind spot. The temporal metric has cells from only one dataset and awaits testing on more data.

\subsection{RQ5: Blind Spots and Ranking Stability Across Architectures}

To test whether the three blind spots are specific to the single-stage detection-based YOLO pair, we evaluate two top-down transformer models of different capacity---ViTPose-B and ViTPose-L---under the detector-free GT-bbox oracle protocol (Section~4). To make the cross-family comparison a same-protocol statistical test, the two YOLO models are additionally evaluated under the oracle protocol (GT-bbox crop, letterbox $640\times640$, no flip; the crop-induced scale offset of the single-stage models is noted where relevant). All five oracle-protocol runs emit one prediction per GT annotation, so no detector error enters any of the comparisons in this section; the detection-based round5-3 column is retained in Tables~\ref{tab:m1x} and \ref{tab:m2x} as the bridge to the RQ1--RQ4 numbers.

\begin{table}[t]
\centering
\caption{M1 stratified mAP$^{50-95}$(P) under the detector-free GT-bbox oracle protocol (pycocotools 2.0.11), four models. Bold $=$ ViTPose-L significantly better than ViTPose-B (all nine ranking cells; CIs in Appendix~A, Table~A6). $^{\ast}$ $=$ the marked YOLO-oracle model significantly better than the other YOLO-oracle model (within-family training-data effect under the shared protocol; CIs in Appendix~A, Table~A8). $\Delta_{\mathrm{occ}}$ (v1 layer $-$ full): ViTPose-B $-0.4666$/$-0.2420$/$-0.3470$, ViTPose-L $-0.4698$/$-0.2381$/$-0.3448$, round5-3-oracle $-0.3310$/$-0.2068$/$-0.2276$, YOLO11s-oracle $-0.3440$/$-0.1947$/$-0.2308$ on COCO/CrowdPose/OCHuman. $^{\dagger}$ Detection-based run, different protocol; bridge to Tables~\ref{tab:m1}--\ref{tab:m3} only.}
\label{tab:m1x}
\renewcommand{\arraystretch}{1.15}
\resizebox{\columnwidth}{!}{%
\begin{tabular}{llccccc}
\toprule
\textbf{Dataset} & \textbf{Layer} & \textbf{ViTPose-B} & \textbf{ViTPose-L} & \textbf{r5-3-oracle} & \textbf{YOLO11s-oracle} & \textbf{r5-3$^{\dagger}$} \\
\midrule
COCO val & full & 0.7676 & \textbf{0.7988} & 0.4250 & \textbf{0.4411}$^{\ast}$ & 0.5712 \\
COCO val & $v{=}2$ & 0.7612 & \textbf{0.7903} & 0.4188 & \textbf{0.4347}$^{\ast}$ & 0.5637 \\
COCO val & $v{=}1$ & 0.3010 & \textbf{0.3290} & 0.0940 & 0.0971 & 0.1747 \\
COCO val & $v{=}0$ & 0 & 0 & 0 & 0 & 0 \\
CrowdPose val & full & 0.8360 & \textbf{0.8653} & \textbf{0.5591}$^{\ast}$ & 0.5148 & 0.7626 \\
CrowdPose val & $v{=}2$ & 0.4905 & \textbf{0.5042} & \textbf{0.2783}$^{\ast}$ & 0.2555 & 0.4183 \\
CrowdPose val & $v{=}1$ & 0.5940 & \textbf{0.6272} & \textbf{0.3523}$^{\ast}$ & 0.3201 & 0.5183 \\
CrowdPose val & $v{=}0$ & 0 & 0 & 0 & 0 & 0 \\
OCHuman test & full & 0.5514 & \textbf{0.6083} & \textbf{0.3546}$^{\ast}$ & 0.3044 & 0.4775 \\
OCHuman test & $v{=}2$ & 0.5388 & \textbf{0.5966} & \textbf{0.3403}$^{\ast}$ & 0.2901 & 0.4696 \\
OCHuman test & $v{=}1$ & 0.2044 & \textbf{0.2636} & \textbf{0.1270}$^{\ast}$ & 0.0736 & 0.1819 \\
OCHuman test & $v{=}0$ & 0 & 0 & 0 & 0 & 0 \\
\bottomrule
\end{tabular}}
\end{table}

\begin{table}[t]
\centering
\caption{M2 joint-angle RMSE ($^{\circ}$, instances pooled per side) under the oracle protocol, four models. Significance between the two ViTPose models is assessed at the per-joint level (Appendix~A, Table~A7): of the 24 comparisons, 5 are significant and all favor ViTPose-L; bold marks the pooled cell whose two per-joint comparisons are both significant. Within-YOLO-oracle significance is also per-joint (Appendix~A, Table~A11): none on COCO, one on CrowdPose (left elbow $v{=}1$, $-3.0^{\circ}$), and three on OCHuman (left elbow $v{=}1$ $-9.2^{\circ}$, right elbow $v{=}2$ $-4.1^{\circ}$, right elbow $v{=}1$ $-8.2^{\circ}$), all favoring round5-3-oracle. $^{\dagger}$ Detection-based run; bridge to Table~\ref{tab:m2} only.}
\label{tab:m2x}
\renewcommand{\arraystretch}{1.15}
\resizebox{\columnwidth}{!}{%
\begin{tabular}{llcccc}
\toprule
\textbf{Dataset} & \textbf{Model} & \textbf{Elbow $v{=}2$} & \textbf{Elbow $v{=}1$} & \textbf{Knee $v{=}2$} & \textbf{Knee $v{=}1$} \\
\midrule
COCO val & ViTPose-B & 17.8 & 31.0 & 14.2 & 33.7 \\
COCO val & ViTPose-L & \textbf{14.9} & 28.0 & 12.6 & 28.5 \\
COCO val & round5-3-oracle & 27.0 & 41.8 & 23.6 & 39.1 \\
COCO val & YOLO11s-oracle & 28.0 & 40.2 & 24.5 & 38.3 \\
COCO val & round5-3$^{\dagger}$ & 26.0 & 35.9 & 20.4 & 29.2 \\
CrowdPose val & ViTPose-B & 14.9 & 30.7 & 13.1 & 28.5 \\
CrowdPose val & ViTPose-L & 13.6 & 28.0 & 11.8 & 26.8 \\
CrowdPose val & round5-3-oracle & 27.9 & 33.1 & 24.6 & 32.4 \\
CrowdPose val & YOLO11s-oracle & 28.8 & 35.6 & 27.1 & 35.2 \\
CrowdPose val & round5-3$^{\dagger}$ & 25.9 & 31.6 & 20.6 & 25.9 \\
OCHuman test & ViTPose-B & 36.5 & 55.3 & 21.6 & 36.9 \\
OCHuman test & ViTPose-L & 33.7 & 51.8 & 19.2 & 33.0 \\
OCHuman test & round5-3-oracle & 37.7 & 49.2 & 24.5 & 38.6 \\
OCHuman test & YOLO11s-oracle & 40.6 & 57.9 & 26.9 & 40.5 \\
OCHuman test & round5-3$^{\dagger}$ & 35.6 & 45.7 & 24.3 & 32.0 \\
\bottomrule
\end{tabular}}
\end{table}

\begin{table}[t]
\centering
\caption{M3 temporal metrics under the oracle protocol (PoseTrack18 val; retained track segments: 868 / 885 for ViTPose-B/L, 709 / 651 for round5-3-oracle / YOLO11s-oracle, over 170 videos). Matched-scope AP: 0.5248 / 0.5548 (ViTPose-B/L), 0.3301 / 0.3085 (round5-3-oracle / YOLO11s-oracle; oracle values, not comparable to the detection-based values of Table~\ref{tab:m3}). AP cost $=$ matched-scope AP after smoothing $-$ raw. YOLO-oracle smoothing is summarized in the text (Section~5.5). $^{\dagger}$ Detection-based run (Table~\ref{tab:m3}); bridge only.}
\label{tab:m3x}
\renewcommand{\arraystretch}{1.15}
\resizebox{\columnwidth}{!}{%
\begin{tabular}{llccccc}
\toprule
\textbf{Model} & \textbf{Convention} & \textbf{disp\_mean} & \textbf{jitter\_mean} & \textbf{jump\_0.05} & \textbf{jump\_0.10} & \textbf{AP cost} \\
\midrule
ViTPose-B & raw & 0.0389 & 0.0526 & 0.1648 & 0.0597 & --- \\
ViTPose-B & pixel-domain smoothing & 0.0370 & 0.0517 & 0.1555 & 0.0557 & $-0.0001$ \\
ViTPose-B & q-domain smoothing & 0.0755 & 0.0549 & 0.2004 & 0.0776 & $-0.0146$ \\
ViTPose-L & raw & 0.0385 & 0.0538 & 0.1599 & 0.0563 & --- \\
ViTPose-L & pixel-domain smoothing & 0.0368 & 0.0530 & 0.1514 & 0.0529 & $-0.0003$ \\
ViTPose-L & q-domain smoothing & 0.0742 & 0.0560 & 0.1966 & 0.0751 & $-0.0185$ \\
round5-3-oracle & raw & 0.0441 & 0.0534 & 0.2229 & 0.0773 & --- \\
YOLO11s-oracle & raw & 0.0463 & 0.0542 & 0.2322 & 0.0828 & --- \\
\bottomrule
\end{tabular}}
\end{table}

Four observations. First, \textbf{the three blind spots persist---and in part deepen---on stronger transformer models, and now on the YOLO pair under the same protocol}. $\Delta_{\mathrm{occ}}$ is negative in every oracle combination: $-0.24$ to $-0.47$ for the transformer pair and $-0.19$ to $-0.34$ for the YOLO-oracle pair, with the transformer magnitudes comparable to or larger than those of the detection-based pair ($-0.24$ to $-0.41$); on COCO the ViTPose oracle $\Delta_{\mathrm{occ}}$ ($-0.47$) even exceeds the detection-based $-0.40$. A stronger architecture or larger capacity does not buy a smaller blind spot: under both protocols, full mAP overstates occluded-scene performance by 19--47 points.

Second, \textbf{the oracle protocol decomposes temporal jitter into detector, training-data, and cross-family components}. Detector removal does not eliminate the jitter in either architecture family: the transformer pair still jumps at 0.165 (B) / 0.160 (L) with no detector in the loop, and the YOLO pair re-run under the oracle protocol jumps at 0.223/0.232, no lower than its detection-based 0.206/0.215 (the oracle runs use different preprocessing and matched segment populations, so this last comparison is indicative rather than a controlled decomposition). The residual high-frequency jitter therefore persists without a detector in the loop; whether it is pose-head-intrinsic or an artifact of the residual preprocessing differences cannot be separated under the current oracle design. Under the shared oracle protocol, all four cross-family comparisons are significant for disp\_mean, jump\_0.05, and jump\_0.10, and three of four are significant for disp\_p95 (the ViTPose-B $-$ round5-3-oracle disp\_p95 CI [$-$0.0275, $+0.0006]$ includes 0; e.g., ViTPose-B $-$ round5-3-oracle jump\_0.05 $-$0.058 [$-$0.080, $-$0.036]; Appendix~A, Table~A10): the transformer family is significantly more temporally stable---a cross-family difference in which architecture is confounded with preprocessing, input resolution, and flip test. The within-YOLO oracle pair differs in no M3 metric (all seven CIs include 0): training-data differences do not change temporal stability, while the cross-family gap does. jitter\_mean stays nearly constant across all six runs and both protocols (0.0521--0.0542), suggesting that the median-filter residual measures a stable property of per-frame heatmap inference rather than a model-specific one. The smoothing-domain finding replicates on the YOLO-oracle pair as well: pixel-domain smoothing lowers the jump rate by 10.7\% (round5-3-oracle) at $+0.0009$ matched-scope AP, while q-domain smoothing raises it by 5.0\% and costs 0.0068 AP. The pixel-domain primary convention therefore generalizes across architectures, and the q-domain failure mechanism is independent of architecture.

Third, \textbf{ranking is stable within both families}. ViTPose-L is significantly better than ViTPose-B in all nine M1 ranking cells (CIs in Appendix~A, Table~A6; e.g., COCO full $+0.0312$ [$+0.0247$, $+0.0329$]), and directionally better in M2 (5 of 24 per-joint comparisons significant, all favoring ViTPose-L, concentrated in the $v{=}2$ layer). Within the YOLO family, the oracle protocol preserves the detection-based ranking on every dataset: YOLO11s-oracle is significantly better on COCO full and $v{=}2$ (round5-3-oracle $\Delta=-0.0161$ [$-$0.0220, $-$0.0099]), and round5-3-oracle is significantly better on CrowdPose and OCHuman in all three layers ($+0.0442$ [$+0.0361$, $+0.0527$] and $+0.0501$ [$+0.0383$, $+0.0633$] full; Table~A8). The domain gap shrinks under the oracle---OCHuman full $+0.1396$ detection-based versus $+0.0501$ oracle---showing that part of round5-3's detection-based advantage is detector-borne, but the training-data effect itself survives detector removal.

Fourth, \textbf{the protocol separates training-data effects from model capacity, now statistically}. On OCHuman's $v{=}1$ layer under the shared protocol, the small occlusion-domain-trained round5-3-oracle achieves left-elbow $47.4^{\circ}$ / left-knee $37.1^{\circ}$ versus ViTPose-B's $54.7^{\circ}$ / $36.7^{\circ}$ (per-side RMSE, consistent with the per-side CIs of Table~A12): round5-3-oracle is significantly more accurate than ViTPose-B in the occluded elbow layer (left elbow, ViTPose-B $-$ round5-3-oracle $\Delta=+7.4^{\circ}$ [$+2.4$, $+12.5$]; Table~A12), and ViTPose-L does not significantly beat it on any $v{=}1$ joint (left elbow $+4.6^{\circ}$ [$-$0.4, $+9.5$] for ViTPose-L, ns; knees ns)---despite ViTPose-L's full-mAP lead of $+0.2538$ (0.6083 vs.\ 0.3546). A capacity advantage without occlusion-domain training data does not translate into occluded-layer angle accuracy; this separation is invisible in full mAP under either protocol and is exactly the information the protocol adds.

Taken together, the same-protocol comparisons show that (i) the three blind spots persist across architecture families under one shared protocol; (ii) within-family training-data effects survive detector removal with compressed magnitude; (iii) temporal stability differs across the two families but not across training data; and (iv) the occlusion-domain-trained smaller model retains a significant occluded-elbow angle advantage over ViTPose-B that even ViTPose-L does not significantly reverse.

\section{Discussion}

\textbf{The magnitude of the three blind spots is quantified, to our knowledge for the first time, under one protocol.} The consistent observation across four datasets and two models is: full mAP's overestimation of occluded-scene performance ($\Delta_{\mathrm{occ}}$ $-0.24$ to $-0.41$), occlusion's amplification of angle error ($v{=}1$-layer elbow RMSE up to $+18.1^{\circ}$), and the high-frequency jitter of tracking-free per-frame outputs (jump rate $\approx0.21$ for both models on the PoseTrack18 val retained-segment subpopulation at the default threshold $\tau_j=0.05$, reducible by about 10\% at zero matched-scope AP cost with pixel-domain smoothing). Together these numbers show that a model selection decision driven solely by OKS-mAP would misread the composition of the gap between the two models tested: on COCO, YOLO11s-pose's significant aggregate advantage comes entirely from visible keypoints and does not extend to the occluded layer, and the between-model difference in occlusion masking ($\Delta\Delta_{\mathrm{occ}}$, Section~5.1) is significant on all three datasets---differences invisible in the aggregate score. The magnitude of these blind spots is not specific to one architecture: under the oracle protocol the transformer pair reproduces the same pattern with even larger occlusion masking, and the YOLO pair re-run under the same protocol reproduces it again with compressed magnitude (Section~5.5); the oracle decomposition shows the remaining temporal jitter to persist after detector removal and the transformer family's jump rate to be significantly lower than the YOLO family's.

\textbf{The boundary between the angle protocol and validity studies.} Our angle metrics use ground-truth-derived angles as the reference and answer ``which model's joint angles are closer to the annotated truth''---a model-comparison tool. They do not validate ``do estimated angles equal a clinical gold standard'', which is the task of the markerless validity literature~\cite{4,5,6} and requires markers or expert annotations. The two are complementary: validity studies certify that a specific model is trustworthy in a specific scenario, while our protocol gives any new model comparable angle-dimension scores on standard datasets. Likewise, the knee-valgus detection consistency ($\kappa$ up to 0.76 at the default threshold $\tau_\varphi{=}15^{\circ}$) asserts only ``consistency with GT annotations'', not clinical meaning.

\textbf{Engineering implications of the temporal metrics.} M3's two findings directly guide system design: first, data-side interventions (occlusion training) do not change temporal stability, so temporal quality must be handled separately; second, the choice of smoothing domain decides the benefit---normalized-domain filtering looks theoretically correct but fails in practice due to the re-injection of GT-box scale noise; third, the oracle decomposition (Section~5.5) shows the residual jitter to persist after detector removal, so better detection is not the lever---post-processing or pose-head-level changes are; separating pose-head-intrinsic from preprocessing contributions requires a preprocessing-matched control experiment. The protocol fixes the pixel domain as the primary convention and recommends that any temporal smoothing method report the joint ``jitter before/after + AP cost'' table.

\textbf{Limitations.} (1) The baseline covers two architecture families under two protocols: within-family comparisons are detection-based for the YOLO pair and oracle-based for the ViTPose pair, while all cross-family comparisons are made under the shared oracle protocol. The YOLO-oracle variant uses a GT-bbox crop with letterboxing rather than the transformers' UDP preprocessing, and the two differ additionally in input resolution ($640\times640$ vs.\ $256\times192$) and in the use of flip test (absent vs.\ present). The cross-family comparisons are therefore explicitly confounded: preprocessing, resolution, and flip-test differences are not separated from architecture differences, and any cross-family gap should be read as a joint effect rather than a pure architecture effect (Section~5.5 uses this phrasing). Extending the baseline to other architectures (e.g., RTMPose, DWPose) is left to future work; the protocol's value is its conventions and tooling. (2) CrowdPose's official visibility clustering (nearly all visible points marked $v{=}2$) together with the full-person recall denominator systematically underestimates its v2-layer AP; on that dataset, cross-layer comparisons use $\Delta_{\mathrm{occ}}$; (3) Angles are 2D image-plane angles affected by viewpoint; they are not 3D joint angles or clinical angles. (4) M3 covers a single dataset, and its retained-segment subpopulation covers only about 53\% of annotated frames; the generality of the temporal conclusions awaits extension. (5) RQ4's correlation analysis has small cell counts ($n=6$ cells over only 3 independent datasets) and is exploratory. (6) COCO's $v{=}1$ layer is only 7.8\% of keypoints, so its $v{=}1$-layer estimates have larger variance (wider CIs). (7) M2's comparisons use each model's own valid-instance set; recomputing on the shared intersection leaves all significance calls unchanged, but models with substantially different coverage should use the intersection variant. Every table states its evaluation convention: M1/M2 numbers follow pycocotools 2.0.11, while PoseTrack's official-metric numbers (Section~4) are reported under the ultralytics internal-evaluator convention because PoseTrack has no official pycocotools reference.

\section{Conclusion}

We proposed and executed a unified supplementary evaluation protocol for 2D human pose estimation, targeting occlusion, joint angle, and temporal stability, together with an open-source harness and a reproducibility contract. Baseline results on four models from two architecture families and four public datasets show that, in every model--dataset combination tested, full OKS-mAP masks occlusion-specific accuracy differences ($\Delta_{\mathrm{occ}}$ up to $-0.47$), that the angle-error protocol captures partially independent information beyond OKS, that, on the matched retained-segment subpopulation of PoseTrack18 val, pixel-domain smoothing reduces temporal jitter at zero cost on the matched-scope AP, and that the detector-free oracle protocol shows the residual jitter to survive detector removal (it is not detector-intrinsic)---with the three blind spots persisting across architectures under a shared oracle protocol, and with cross-family ranking, training-data, and temporal-stability effects separable at significance level. The protocol is positioned as a complement to, not a replacement for, OKS-mAP: it does not change the underlying convention of mainstream rankings, but adds three previously invisible diagnostic dimensions for model selection and system design. The code and all raw predictions are open-sourced.


\appendices
\renewcommand{\thetable}{A\arabic{table}}
\setcounter{table}{0}
\section{Bootstrap Confidence Intervals}

All intervals are per-comparison 95\% percentile bootstrap CIs from 10,000 resamples with seed 42, without multiple-comparison correction; significance serves only as a secondary aid, and the core conclusions rest on three-dataset directional consistency and effect sizes. A Benjamini--Hochberg FDR sensitivity analysis over the core comparison families is in Table~A14. The resampling unit is the image for M1 and M2 (the unit over which each metric is aggregated before the model comparison) and the track segment for M3. Paired comparisons share the same resample index across the two quantities; between-model comparisons resample the two models' units independently. Sign convention: $\Delta=$ YOLO11s-pose $-$ round5-3, so $\Delta<0$ favors round5-3 for AP and $\kappa$, and favors YOLO11s-pose for RMSE; for the oracle-protocol comparisons (Tables~A6--A7), $\Delta=$ ViTPose-L $-$ ViTPose-B, so $\Delta>0$ favors ViTPose-L for AP and $\Delta<0$ favors ViTPose-L for RMSE; for Table~A8, $\Delta=$ round5-3-oracle $-$ YOLO11s-oracle, so $\Delta>0$ favors round5-3-oracle for AP and $\Delta<0$ favors it for RMSE; for Table~A9, $\Delta=$ (transformer model) $-$ (YOLO-oracle model) as indicated by the column headers, so $\Delta>0$ favors the first-named model for AP and $\Delta<0$ favors it for RMSE; for Table~A10, $\Delta=$ (transformer model) $-$ (YOLO-oracle model) per row, so $\Delta<0$ favors the transformer (lower jitter); for Tables~A11--A12, $\Delta=$ (first-named model) $-$ (second-named model) per the table header, so $\Delta<0$ favors the first-named model for RMSE; for Table~A13, $\Delta=$ RMSE$(v{=}1)-$RMSE$(v{=}2)$ within the same model, so $\Delta>0$ means the occluded layer has higher error; for Table~A14, $\Delta=$ round5-3 $-$ YOLO11s-pose within each family, so $\Delta>0$ favors round5-3 for AP and favors YOLO11s-pose for RMSE and jitter metrics. Table~A15 reports raw MAE/P95 point estimates without intervals. $^{*}$ marks intervals that exclude 0. Every entry of the fixed comparison list is tabulated; none is selected.


\begin{table}[t]
\centering
\caption{M1 per-layer AP comparison with bootstrap CIs. $\Delta=$ YOLO11s-pose $-$ round5-3; $^{*}$ CI excludes 0.}
\label{tab:a1}
\renewcommand{\arraystretch}{1.15}
\resizebox{\columnwidth}{!}{%
\begin{tabular}{llccc}
\toprule
\textbf{Dataset} & \textbf{Layer} & \textbf{AP r5-3} & \textbf{AP YOLO} & \textbf{$\Delta$ AP [95\% CI]} \\
\midrule
COCO val & full & 0.5712 & 0.5816 & 0.0104 [0.0059, 0.0155]$^{*}$ \\
COCO val & $v{=}2$ & 0.5637 & 0.5728 & 0.0091 [0.0048, 0.0144]$^{*}$ \\
COCO val & $v{=}1$ & 0.1747 & 0.1719 & -0.0028 [-0.0070, 0.0022] \\
COCO val & $v{=}0$ & 0.0000 & 0.0000 & 0.0000 [0.0000, 0.0000] \\
CrowdPose val & full & 0.7626 & 0.6938 & -0.0687 [-0.0754, -0.0616]$^{*}$ \\
CrowdPose val & $v{=}2$ & 0.4183 & 0.3768 & -0.0415 [-0.0465, -0.0362]$^{*}$ \\
CrowdPose val & $v{=}1$ & 0.5183 & 0.4573 & -0.0610 [-0.0677, -0.0540]$^{*}$ \\
CrowdPose val & $v{=}0$ & 0.0000 & 0.0000 & 0.0000 [0.0000, 0.0000] \\
OCHuman test & full & 0.4775 & 0.3379 & -0.1396 [-0.1508, -0.1281]$^{*}$ \\
OCHuman test & $v{=}2$ & 0.4696 & 0.3314 & -0.1383 [-0.1488, -0.1261]$^{*}$ \\
OCHuman test & $v{=}1$ & 0.1819 & 0.0765 & -0.1055 [-0.1153, -0.0973]$^{*}$ \\
OCHuman test & $v{=}0$ & 0.0000 & 0.0000 & 0.0000 [0.0000, 0.0000] \\
\bottomrule
\end{tabular}
}
\vspace{0.4em}\footnotesize Unit: image (both models share the same image universe; paired resampling). 10,000 resamples, seed 42, 95\% percentile CI.
\end{table}

\begin{table}[t]
\centering
\caption{M1 $\Delta_{\mathrm{occ}}$ (mAP($v{=}1$) $-$ mAP(full)) per model, and the model difference, with bootstrap CIs.}
\label{tab:a2}
\renewcommand{\arraystretch}{1.15}
\resizebox{\columnwidth}{!}{%
\begin{tabular}{lccc}
\toprule
\textbf{Dataset} & $\Delta_{\mathrm{occ}}$ r5-3 [95\% CI] & $\Delta_{\mathrm{occ}}$ YOLO [95\% CI] & $\Delta\Delta_{\mathrm{occ}}$ [95\% CI] \\
\midrule
COCO val & -0.3965 [-0.4075, -0.3858] & -0.4097 [-0.4205, -0.3989] & -0.0132 [-0.0177, -0.0086]$^{*}$ \\
CrowdPose val & -0.2443 [-0.2565, -0.2322] & -0.2366 [-0.2488, -0.2246] & 0.0077 [0.0041, 0.0110]$^{*}$ \\
OCHuman test & -0.2955 [-0.3080, -0.2842] & -0.2614 [-0.2729, -0.2527] & 0.0341 [0.0237, 0.0429]$^{*}$ \\
\bottomrule
\end{tabular}
}
\vspace{0.4em}\footnotesize Unit: image (paired); $\Delta\Delta_{\mathrm{occ}}$ is computed element-wise on the paired resample. 10,000 resamples, seed 42, 95\% percentile CI.
\end{table}

\begin{table}[t]
\centering
\caption{M2 joint-angle RMSE ($^{\circ}$, instances pooled per side) with bootstrap CIs. $\Delta=$ YOLO11s-pose $-$ round5-3 (negative $=$ YOLO lower error); $^{*}$ CI excludes 0.}
\label{tab:a3}
\renewcommand{\arraystretch}{1.15}
\resizebox{\columnwidth}{!}{%
\begin{tabular}{lllcccc}
\toprule
\textbf{Dataset} & \textbf{Joint} & \textbf{Layer} & \textbf{RMSE r5-3} & \textbf{RMSE YOLO} & \textbf{$\Delta$ RMSE [95\% CI]} & \textbf{$n_{\mathrm{img}}$} \\
\midrule
COCO val & Left elbow & $v{=}2$ & 27.0 & 26.3 & -0.7 [-3.1, 1.7] & 1720 / 1727 \\
COCO val & Left elbow & $v{=}1$ & 33.5 & 36.6 & 3.1 [-4.8, 10.9] & 248 / 247 \\
COCO val & Right elbow & $v{=}2$ & 24.9 & 24.5 & -0.5 [-2.8, 1.9] & 1732 / 1730 \\
COCO val & Right elbow & $v{=}1$ & 37.7 & 34.8 & -2.9 [-9.1, 3.5] & 290 / 292 \\
COCO val & Left knee & $v{=}2$ & 20.3 & 20.9 & 0.6 [-2.0, 3.1] & 1320 / 1325 \\
COCO val & Left knee & $v{=}1$ & 30.8 & 32.2 & 1.4 [-8.9, 11.3] & 180 / 182 \\
COCO val & Right knee & $v{=}2$ & 20.5 & 21.2 & 0.7 [-1.6, 3.2] & 1302 / 1306 \\
COCO val & Right knee & $v{=}1$ & 27.7 & 30.3 & 2.6 [-6.9, 12.3] & 202 / 205 \\
CrowdPose val & Left elbow & $v{=}2$ & 27.5 & 27.3 & -0.2 [-2.9, 2.6] & 790 / 793 \\
CrowdPose val & Left elbow & $v{=}1$ & 31.3 & 35.0 & 3.7 [1.4, 6.1]$^{*}$ & 1319 / 1321 \\
CrowdPose val & Right elbow & $v{=}2$ & 24.1 & 24.6 & 0.6 [-2.0, 3.2] & 810 / 811 \\
CrowdPose val & Right elbow & $v{=}1$ & 31.9 & 35.0 & 3.1 [0.7, 5.5]$^{*}$ & 1343 / 1338 \\
CrowdPose val & Left knee & $v{=}2$ & 21.1 & 21.3 & 0.2 [-3.3, 3.7] & 575 / 575 \\
CrowdPose val & Left knee & $v{=}1$ & 25.7 & 30.5 & 4.9 [2.1, 7.7]$^{*}$ & 1085 / 1081 \\
CrowdPose val & Right knee & $v{=}2$ & 20.0 & 21.2 & 1.2 [-2.0, 4.4] & 570 / 571 \\
CrowdPose val & Right knee & $v{=}1$ & 26.2 & 30.5 & 4.4 [1.4, 7.2]$^{*}$ & 1104 / 1096 \\
OCHuman test & Left elbow & $v{=}2$ & 35.6 & 39.7 & 4.2 [1.7, 6.6]$^{*}$ & 1931 / 1951 \\
OCHuman test & Left elbow & $v{=}1$ & 45.4 & 56.2 & 10.8 [7.0, 14.7]$^{*}$ & 860 / 871 \\
OCHuman test & Right elbow & $v{=}2$ & 35.6 & 39.5 & 3.9 [1.4, 6.4]$^{*}$ & 1935 / 1947 \\
OCHuman test & Right elbow & $v{=}1$ & 46.0 & 59.2 & 13.2 [9.2, 17.3]$^{*}$ & 822 / 831 \\
OCHuman test & Left knee & $v{=}2$ & 24.4 & 28.0 & 3.6 [1.1, 6.2]$^{*}$ & 1507 / 1514 \\
OCHuman test & Left knee & $v{=}1$ & 30.7 & 39.5 & 8.8 [3.7, 14.0]$^{*}$ & 498 / 499 \\
OCHuman test & Right knee & $v{=}2$ & 24.2 & 28.5 & 4.3 [1.8, 6.9]$^{*}$ & 1499 / 1503 \\
OCHuman test & Right knee & $v{=}1$ & 33.3 & 39.9 & 6.6 [0.9, 12.4]$^{*}$ & 525 / 534 \\
\bottomrule
\end{tabular}
}
\vspace{0.4em}\footnotesize Unit: image, resampled independently per model (each model's image set contains images with at least one valid instance of that joint and layer); last column $n_{\mathrm{img}}$: round5-3 / YOLO11s-pose. 10,000 resamples, seed 42, 95\% percentile CI.
\end{table}

\begin{table}[t]
\centering
\caption{Knee-valgus detection consistency ($\kappa$, sensitivity, specificity) with bootstrap CIs. $\Delta=$ YOLO11s-pose $-$ round5-3; $^{*}$ CI excludes 0.}
\label{tab:a4}
\renewcommand{\arraystretch}{1.15}
\resizebox{\columnwidth}{!}{%
\begin{tabular}{llccc cc}
\toprule
\textbf{Dataset} & \textbf{$\tau_\varphi$} & \textbf{$\kappa$ r5-3 [95\% CI]} & \textbf{$\kappa$ YOLO [95\% CI]} & \multicolumn{3}{c}{\textbf{Model difference [95\% CI]}} \\
 & & & & $\Delta\kappa$ & $\Delta$ sens. & $\Delta$ spec. \\
\midrule
COCO val & $10^{\circ}$ & 0.724 [0.702, 0.745] & 0.697 [0.673, 0.719] & -0.027 [-0.058, 0.003] & -0.015 [-0.043, 0.014] & -0.012 [-0.029, 0.005] \\
COCO val & $15^{\circ}$ & 0.759 [0.738, 0.780] & 0.745 [0.723, 0.767] & -0.014 [-0.045, 0.016] & -0.006 [-0.039, 0.026] & -0.006 [-0.019, 0.006] \\
COCO val & $17^{\circ}$ & 0.773 [0.751, 0.794] & 0.759 [0.737, 0.780] & -0.014 [-0.045, 0.017] & -0.006 [-0.039, 0.027] & -0.006 [-0.017, 0.006] \\
CrowdPose val & $10^{\circ}$ & 0.711 [0.692, 0.729] & 0.666 [0.646, 0.685] & -0.045 [-0.073, -0.018]$^{*}$ & -0.029 [-0.052, -0.004] & -0.017 [-0.032, -0.001] \\
CrowdPose val & $15^{\circ}$ & 0.739 [0.719, 0.757] & 0.669 [0.648, 0.690] & -0.069 [-0.098, -0.041]$^{*}$ & -0.063 [-0.091, -0.034] & -0.011 [-0.024, 0.002] \\
CrowdPose val & $17^{\circ}$ & 0.731 [0.711, 0.750] & 0.665 [0.643, 0.687] & -0.067 [-0.096, -0.037]$^{*}$ & -0.067 [-0.097, -0.037] & -0.007 [-0.019, 0.005] \\
OCHuman test & $10^{\circ}$ & 0.601 [0.577, 0.625] & 0.528 [0.502, 0.554] & -0.073 [-0.108, -0.037]$^{*}$ & -0.041 [-0.072, -0.010] & -0.030 [-0.051, -0.010] \\
OCHuman test & $15^{\circ}$ & 0.602 [0.576, 0.627] & 0.519 [0.490, 0.546] & -0.084 [-0.121, -0.046]$^{*}$ & -0.054 [-0.090, -0.017] & -0.026 [-0.041, -0.011] \\
OCHuman test & $17^{\circ}$ & 0.603 [0.576, 0.628] & 0.501 [0.473, 0.530] & -0.101 [-0.140, -0.062]$^{*}$ & -0.071 [-0.110, -0.032] & -0.026 [-0.039, -0.011] \\
\bottomrule
\end{tabular}
}
\vspace{0.4em}\footnotesize Unit: image (binarized $\phi$ instances pooled per image), resampled independently per model. 10,000 resamples, seed 42, 95\% percentile CI.
\end{table}

\begin{table}[t]
\centering
\caption{M3 temporal metrics: per-domain point values, within-model domain differences (px$-$raw, q$-$raw), and between-model difference on the raw domain, all with bootstrap CIs. $\Delta$ (model) $=$ YOLO11s-pose $-$ round5-3; $^{*}$ CI excludes 0.}
\label{tab:a5}
\renewcommand{\arraystretch}{1.15}
\resizebox{\columnwidth}{!}{%
\begin{tabular}{lccccccc}
\toprule
\textbf{Metric} & \textbf{raw r5-3} & \textbf{raw YOLO} & \textbf{$\Delta$ raw} & \textbf{px$-$raw r5-3} & \textbf{px$-$raw YOLO} & \textbf{q$-$raw r5-3} & \textbf{q$-$raw YOLO} \\
\midrule
$\bar d$ & 0.0409 & 0.0430 & 0.0022 [-0.0019, 0.0062] & -0.0029 [-0.0031, -0.0027]$^{*}$ & -0.0031 [-0.0033, -0.0030]$^{*}$ & 0.0023 [0.0015, 0.0031]$^{*}$ & 0.0031 [0.0022, 0.0041]$^{*}$ \\
$\bar j$ & 0.0521 & 0.0535 & 0.0014 [-0.0047, 0.0075] & -0.0012 [-0.0013, -0.0011]$^{*}$ & -0.0013 [-0.0013, -0.0012]$^{*}$ & 0.0007 [0.0003, 0.0011]$^{*}$ & 0.0010 [0.0006, 0.0015]$^{*}$ \\
$d_{95}$ & 0.1194 & 0.1246 & 0.0051 [-0.0081, 0.0183] & -0.0086 [-0.0093, -0.0076]$^{*}$ & -0.0088 [-0.0096, -0.0080]$^{*}$ & 0.0079 [0.0050, 0.0111]$^{*}$ & 0.0103 [0.0070, 0.0140]$^{*}$ \\
$j_{95}$ & 0.1669 & 0.1719 & 0.0050 [-0.0177, 0.0275] & -0.0038 [-0.0046, -0.0032]$^{*}$ & -0.0044 [-0.0049, -0.0035]$^{*}$ & 0.0031 [0.0008, 0.0056]$^{*}$ & 0.0043 [0.0021, 0.0073]$^{*}$ \\
jump$_{0.05}$ & 0.2062 & 0.2153 & 0.0091 [-0.0145, 0.0326] & -0.0191 [-0.0206, -0.0177]$^{*}$ & -0.0210 [-0.0224, -0.0195]$^{*}$ & 0.0139 [0.0091, 0.0188]$^{*}$ & 0.0183 [0.0132, 0.0235]$^{*}$ \\
jump$_{0.10}$ & 0.0726 & 0.0777 & 0.0051 [-0.0086, 0.0190] & -0.0092 [-0.0102, -0.0082]$^{*}$ & -0.0098 [-0.0108, -0.0090]$^{*}$ & 0.0081 [0.0052, 0.0113]$^{*}$ & 0.0100 [0.0067, 0.0135]$^{*}$ \\
jump$_{0.20}$ & 0.0152 & 0.0184 & 0.0032 [-0.0015, 0.0079] & -0.0025 [-0.0030, -0.0021]$^{*}$ & -0.0030 [-0.0035, -0.0026]$^{*}$ & 0.0024 [0.0011, 0.0038]$^{*}$ & 0.0030 [0.0014, 0.0047]$^{*}$ \\
\bottomrule
\end{tabular}
}
\vspace{0.4em}\footnotesize Unit: track segment. Domain differences are paired (same resample index across domains, segments fixed by the raw matching); between-model differences resample each model's segments independently. 10,000 resamples, seed 42 (1042 for the YOLO11s-pose between-model distribution), 95\% percentile CI.
\end{table}

\begin{table}[t]
\centering
\caption{M1 per-layer AP comparison under the detector-free oracle protocol with bootstrap CIs. $\Delta=$ ViTPose-L $-$ ViTPose-B; $^{*}$ CI excludes 0.}
\label{tab:a6}
\renewcommand{\arraystretch}{1.15}
\resizebox{\columnwidth}{!}{%
\begin{tabular}{llccc}
\toprule
\textbf{Dataset} & \textbf{Layer} & \textbf{AP ViTPose-B} & \textbf{AP ViTPose-L} & \textbf{$\Delta$ AP [95\% CI]} \\
\midrule
COCO val & full & 0.7676 & 0.7988 & 0.0312 [0.0247, 0.0329]$^{*}$ \\
COCO val & $v{=}2$ & 0.7612 & 0.7903 & 0.0291 [0.0246, 0.0329]$^{*}$ \\
COCO val & $v{=}1$ & 0.3010 & 0.3290 & 0.0280 [0.0241, 0.0333]$^{*}$ \\
COCO val & $v{=}0$ & 0.0000 & 0.0000 & 0.0000 [0.0000, 0.0000] \\
CrowdPose val & full & 0.8360 & 0.8653 & 0.0293 [0.0238, 0.0320]$^{*}$ \\
CrowdPose val & $v{=}2$ & 0.4905 & 0.5042 & 0.0137 [0.0113, 0.0179]$^{*}$ \\
CrowdPose val & $v{=}1$ & 0.5940 & 0.6272 & 0.0332 [0.0294, 0.0382]$^{*}$ \\
CrowdPose val & $v{=}0$ & 0.0000 & 0.0000 & 0.0000 [0.0000, 0.0000] \\
OCHuman test & full & 0.5514 & 0.6083 & 0.0570 [0.0477, 0.0658]$^{*}$ \\
OCHuman test & $v{=}2$ & 0.5388 & 0.5966 & 0.0578 [0.0493, 0.0677]$^{*}$ \\
OCHuman test & $v{=}1$ & 0.2044 & 0.2636 & 0.0592 [0.0511, 0.0678]$^{*}$ \\
OCHuman test & $v{=}0$ & 0.0000 & 0.0000 & 0.0000 [0.0000, 0.0000] \\
\bottomrule
\end{tabular}
}
\vspace{0.4em}\footnotesize Unit: image (paired; both models share the same image universe under the oracle protocol). 10,000 resamples, seed 42, 95\% percentile CI.
\end{table}

\begin{table}[t]
\centering
\caption{M2 joint-angle RMSE ($^{\circ}$, per side) under the detector-free oracle protocol with bootstrap CIs. $\Delta=$ ViTPose-L $-$ ViTPose-B (negative $=$ ViTPose-L lower error); $^{*}$ CI excludes 0.}
\label{tab:a7}
\renewcommand{\arraystretch}{1.15}
\resizebox{\columnwidth}{!}{%
\begin{tabular}{lllcccc}
\toprule
\textbf{Dataset} & \textbf{Joint} & \textbf{Layer} & \textbf{RMSE ViTPose-B} & \textbf{RMSE ViTPose-L} & \textbf{$\Delta$ RMSE [95\% CI]} & \textbf{$n_{\mathrm{img}}$} \\
\midrule
COCO val & Left elbow & $v{=}2$ & 18.3 & 15.5 & -2.7 [-5.1, -0.4]$^{*}$ & 1747 / 1750 \\
COCO val & Left elbow & $v{=}1$ & 29.0 & 27.8 & -1.2 [-9.2, 6.7] & 261 / 262 \\
COCO val & Right elbow & $v{=}2$ & 17.3 & 14.3 & -3.0 [-5.5, -0.5]$^{*}$ & 1757 / 1756 \\
COCO val & Right elbow & $v{=}1$ & 32.6 & 28.2 & -4.4 [-12.4, 3.5] & 307 / 312 \\
COCO val & Left knee & $v{=}2$ & 13.9 & 12.9 & -0.9 [-3.3, 1.5] & 1347 / 1349 \\
COCO val & Left knee & $v{=}1$ & 33.4 & 25.2 & -8.2 [-17.4, 1.1] & 187 / 188 \\
COCO val & Right knee & $v{=}2$ & 14.5 & 12.3 & -2.1 [-4.3, 0.1] & 1328 / 1329 \\
COCO val & Right knee & $v{=}1$ & 33.9 & 31.1 & -2.8 [-13.2, 7.6] & 212 / 214 \\
CrowdPose val & Left elbow & $v{=}2$ & 14.9 & 14.1 & -0.8 [-3.0, 1.5] & 795 / 793 \\
CrowdPose val & Left elbow & $v{=}1$ & 31.5 & 29.5 & -2.0 [-4.6, 0.6] & 1327 / 1327 \\
CrowdPose val & Right elbow & $v{=}2$ & 14.8 & 13.0 & -1.8 [-4.0, 0.3] & 811 / 811 \\
CrowdPose val & Right elbow & $v{=}1$ & 29.8 & 26.5 & -3.3 [-5.9, -0.6]$^{*}$ & 1343 / 1344 \\
CrowdPose val & Left knee & $v{=}2$ & 13.0 & 12.1 & -0.9 [-3.9, 2.1] & 576 / 577 \\
CrowdPose val & Left knee & $v{=}1$ & 28.1 & 27.3 & -0.8 [-4.7, 3.0] & 1081 / 1086 \\
CrowdPose val & Right knee & $v{=}2$ & 13.2 & 11.5 & -1.8 [-4.6, 1.0] & 571 / 571 \\
CrowdPose val & Right knee & $v{=}1$ & 28.9 & 26.4 & -2.6 [-6.6, 1.6] & 1095 / 1099 \\
OCHuman test & Left elbow & $v{=}2$ & 37.2 & 33.9 & -3.2 [-6.2, -0.3]$^{*}$ & 1848 / 1850 \\
OCHuman test & Left elbow & $v{=}1$ & 54.7 & 52.0 & -2.8 [-7.4, 1.7] & 690 / 708 \\
OCHuman test & Right elbow & $v{=}2$ & 35.9 & 33.4 & -2.5 [-5.6, 0.5] & 1855 / 1849 \\
OCHuman test & Right elbow & $v{=}1$ & 56.0 & 51.6 & -4.4 [-9.2, 0.5] & 659 / 687 \\
OCHuman test & Left knee & $v{=}2$ & 21.0 & 18.0 & -3.0 [-5.8, -0.2]$^{*}$ & 1428 / 1453 \\
OCHuman test & Left knee & $v{=}1$ & 36.7 & 31.1 & -5.7 [-12.2, 0.9] & 358 / 377 \\
OCHuman test & Right knee & $v{=}2$ & 22.3 & 20.4 & -1.9 [-4.9, 1.2] & 1435 / 1449 \\
OCHuman test & Right knee & $v{=}1$ & 37.0 & 34.7 & -2.4 [-9.1, 4.5] & 375 / 402 \\
\bottomrule
\end{tabular}
}
\vspace{0.4em}\footnotesize Unit: image, resampled independently per model (each model's image set contains images with at least one valid instance of that joint and layer); last column $n_{\mathrm{img}}$: ViTPose-B / ViTPose-L. 10,000 resamples, seed 42, 95\% percentile CI.
\end{table}

\begin{table}[t]
\centering
\caption{M1 per-layer AP comparison for the YOLO pair under the detector-free oracle protocol with bootstrap CIs. $\Delta=$ round5-3-oracle $-$ YOLO11s-oracle; $^{*}$ CI excludes 0.}
\label{tab:a8}
\renewcommand{\arraystretch}{1.15}
\resizebox{\columnwidth}{!}{%
\begin{tabular}{llccc}
\toprule
\textbf{Dataset} & \textbf{Layer} & \textbf{AP r5-3-o} & \textbf{AP YOLO11s-o} & \textbf{$\Delta$ AP [95\% CI]} \\
\midrule
COCO val & full & 0.4250 & 0.4411 & -0.0161 [-0.0220, -0.0099]$^{*}$ \\
COCO val & $v{=}2$ & 0.4188 & 0.4347 & -0.0159 [-0.0213, -0.0092]$^{*}$ \\
COCO val & $v{=}1$ & 0.0940 & 0.0971 & -0.0030 [-0.0072, 0.0015] \\
CrowdPose val & full & 0.5591 & 0.5148 & 0.0442 [0.0361, 0.0527]$^{*}$ \\
CrowdPose val & $v{=}2$ & 0.2783 & 0.2555 & 0.0228 [0.0173, 0.0288]$^{*}$ \\
CrowdPose val & $v{=}1$ & 0.3523 & 0.3201 & 0.0322 [0.0249, 0.0399]$^{*}$ \\
OCHuman test & full & 0.3546 & 0.3044 & 0.0501 [0.0383, 0.0633]$^{*}$ \\
OCHuman test & $v{=}2$ & 0.3403 & 0.2901 & 0.0502 [0.0387, 0.0636]$^{*}$ \\
OCHuman test & $v{=}1$ & 0.1270 & 0.0736 & 0.0534 [0.0453, 0.0635]$^{*}$ \\
\bottomrule
\end{tabular}
}
\vspace{0.4em}\footnotesize Unit: image (paired; both models share the same image universe under the oracle protocol). Point estimate is the paired point difference; CI from the paired bootstrap. 10,000 resamples, seed 42, 95\% percentile CI.
\end{table}

\begin{table}[t]
\centering
\caption{M1 cross-family per-layer AP deltas under the detector-free oracle protocol with bootstrap CIs. Each cell is $\Delta=$ transformer model $-$ YOLO-oracle model, so $\Delta>0$ favors the transformer. $^{*}$ CI excludes 0.}
\label{tab:a9}
\renewcommand{\arraystretch}{1.15}
\resizebox{\columnwidth}{!}{%
\begin{tabular}{llcccc}
\toprule
\textbf{Dataset} & \textbf{Layer} & \textbf{B$-$r5-3-o} & \textbf{B$-$YOLO11s-o} & \textbf{L$-$r5-3-o} & \textbf{L$-$YOLO11s-o} \\
\midrule
COCO val & full & 0.3426 [0.3352, 0.3543]$^{*}$ & 0.3265 [0.3197, 0.3379]$^{*}$ & 0.3738 [0.3637, 0.3832]$^{*}$ & 0.3577 [0.3481, 0.3669]$^{*}$ \\
COCO val & $v{=}2$ & 0.3424 [0.3324, 0.3512]$^{*}$ & 0.3265 [0.3175, 0.3355]$^{*}$ & 0.3715 [0.3608, 0.3803]$^{*}$ & 0.3556 [0.3461, 0.3647]$^{*}$ \\
COCO val & $v{=}1$ & 0.2070 [0.1966, 0.2147]$^{*}$ & 0.2039 [0.1937, 0.2119]$^{*}$ & 0.2350 [0.2249, 0.2440]$^{*}$ & 0.2319 [0.2218, 0.2411]$^{*}$ \\
CrowdPose val & full & 0.2769 [0.2653, 0.2884]$^{*}$ & 0.3212 [0.3107, 0.3317]$^{*}$ & 0.3062 [0.2933, 0.3163]$^{*}$ & 0.3505 [0.3384, 0.3596]$^{*}$ \\
CrowdPose val & $v{=}2$ & 0.2122 [0.1990, 0.2241]$^{*}$ & 0.2350 [0.2224, 0.2469]$^{*}$ & 0.2259 [0.2134, 0.2391]$^{*}$ & 0.2487 [0.2367, 0.2617]$^{*}$ \\
CrowdPose val & $v{=}1$ & 0.2417 [0.2318, 0.2519]$^{*}$ & 0.2739 [0.2645, 0.2843]$^{*}$ & 0.2749 [0.2650, 0.2861]$^{*}$ & 0.3072 [0.2977, 0.3183]$^{*}$ \\
OCHuman test & full & 0.1968 [0.1798, 0.2107]$^{*}$ & 0.2469 [0.2317, 0.2604]$^{*}$ & 0.2538 [0.2360, 0.2681]$^{*}$ & 0.3039 [0.2875, 0.3178]$^{*}$ \\
OCHuman test & $v{=}2$ & 0.1985 [0.1821, 0.2132]$^{*}$ & 0.2487 [0.2344, 0.2632]$^{*}$ & 0.2562 [0.2401, 0.2725]$^{*}$ & 0.3065 [0.2918, 0.3227]$^{*}$ \\
OCHuman test & $v{=}1$ & 0.0774 [0.0647, 0.0893]$^{*}$ & 0.1308 [0.1196, 0.1429]$^{*}$ & 0.1366 [0.1228, 0.1497]$^{*}$ & 0.1900 [0.1769, 0.2038]$^{*}$ \\
\bottomrule
\end{tabular}
}
\vspace{0.4em}\footnotesize Unit: image (paired; all four models share the same image universe under the oracle protocol). Point estimate is the paired point difference; CI from the paired bootstrap. 10,000 resamples, seed 42, 95\% percentile CI.
\end{table}

\begin{table}[t]
\centering
\caption{M3 cross-family temporal-metric deltas on PoseTrack18 val under the detector-free oracle protocol with bootstrap CIs. $\Delta=$ transformer model $-$ YOLO-oracle model, so $\Delta<0$ favors the transformer (lower jitter). $^{*}$ CI excludes 0.}
\label{tab:a10}
\renewcommand{\arraystretch}{1.15}
\resizebox{\columnwidth}{!}{%
\begin{tabular}{llc}
\toprule
\textbf{Comparison} & \textbf{Metric} & \textbf{$\Delta$ [95\% CI]} \\
\midrule
ViTPose-B $-$ r5-3-o & $\bar d$ & -0.0052 [-0.0092, -0.0009]$^{*}$ \\
ViTPose-B $-$ r5-3-o & $d_{95}$ & -0.0136 [-0.0275, 0.0006] \\
ViTPose-B $-$ r5-3-o & jump$_{0.05}$ & -0.0581 [-0.0800, -0.0355]$^{*}$ \\
ViTPose-B $-$ r5-3-o & jump$_{0.10}$ & -0.0176 [-0.0306, -0.0045]$^{*}$ \\
ViTPose-B $-$ YOLO11s-o & $\bar d$ & -0.0074 [-0.0118, -0.0030]$^{*}$ \\
ViTPose-B $-$ YOLO11s-o & $d_{95}$ & -0.0188 [-0.0337, -0.0040]$^{*}$ \\
ViTPose-B $-$ YOLO11s-o & jump$_{0.05}$ & -0.0675 [-0.0904, -0.0446]$^{*}$ \\
ViTPose-B $-$ YOLO11s-o & jump$_{0.10}$ & -0.0232 [-0.0372, -0.0097]$^{*}$ \\
ViTPose-L $-$ r5-3-o & $\bar d$ & -0.0056 [-0.0101, -0.0010]$^{*}$ \\
ViTPose-L $-$ r5-3-o & $d_{95}$ & -0.0184 [-0.0323, -0.0045]$^{*}$ \\
ViTPose-L $-$ r5-3-o & jump$_{0.05}$ & -0.0629 [-0.0850, -0.0413]$^{*}$ \\
ViTPose-L $-$ r5-3-o & jump$_{0.10}$ & -0.0209 [-0.0340, -0.0081]$^{*}$ \\
ViTPose-L $-$ YOLO11s-o & $\bar d$ & -0.0078 [-0.0126, -0.0029]$^{*}$ \\
ViTPose-L $-$ YOLO11s-o & $d_{95}$ & -0.0236 [-0.0381, -0.0091]$^{*}$ \\
ViTPose-L $-$ YOLO11s-o & jump$_{0.05}$ & -0.0723 [-0.0958, -0.0492]$^{*}$ \\
ViTPose-L $-$ YOLO11s-o & jump$_{0.10}$ & -0.0266 [-0.0403, -0.0128]$^{*}$ \\
\bottomrule
\end{tabular}
}
\vspace{0.4em}\footnotesize Unit: track segment (paired; matched-scope segments, one prediction per GT annotation under the oracle protocol). 10,000 resamples, seed 42, 95\% percentile CI.
\end{table}

\begin{table}[t]
\centering
\caption{M2 joint-angle RMSE ($^{\circ}$, per side) for the YOLO pair under the detector-free oracle protocol with bootstrap CIs. $\Delta=$ round5-3-oracle $-$ YOLO11s-oracle (negative $=$ round5-3-oracle lower error); $^{*}$ CI excludes 0.}
\label{tab:a11}
\renewcommand{\arraystretch}{1.15}
\resizebox{\columnwidth}{!}{%
\begin{tabular}{lllcccc}
\toprule
\textbf{Dataset} & \textbf{Joint} & \textbf{Layer} & \textbf{RMSE r5-3-o} & \textbf{RMSE YOLO11s-o} & \textbf{$\Delta$ RMSE [95\% CI]} & \textbf{$n_{\mathrm{img}}$} \\
\midrule
COCO val & Left elbow & $v{=}2$ & 27.9 & 29.5 & -1.6 [-4.2, 1.0] & 1635 / 1651 \\
COCO val & Left elbow & $v{=}1$ & 39.6 & 37.0 & 2.5 [-5.8, 11.0] & 209 / 206 \\
COCO val & Right elbow & $v{=}2$ & 26.1 & 26.4 & -0.3 [-2.7, 2.0] & 1649 / 1667 \\
COCO val & Right elbow & $v{=}1$ & 43.6 & 42.6 & 0.9 [-6.4, 8.2] & 247 / 249 \\
COCO val & Left knee & $v{=}2$ & 23.6 & 24.3 & -0.6 [-3.8, 2.5] & 1247 / 1262 \\
COCO val & Left knee & $v{=}1$ & 38.9 & 40.1 & -1.3 [-12.0, 9.2] & 149 / 133 \\
COCO val & Right knee & $v{=}2$ & 23.7 & 24.8 & -1.1 [-4.1, 1.7] & 1225 / 1238 \\
COCO val & Right knee & $v{=}1$ & 39.2 & 36.7 & 2.6 [-5.7, 11.1] & 173 / 160 \\
CrowdPose val & Left elbow & $v{=}2$ & 29.0 & 30.1 & -1.0 [-4.2, 2.1] & 770 / 768 \\
CrowdPose val & Left elbow & $v{=}1$ & 32.8 & 35.8 & -3.0 [-5.5, -0.4]$^{*}$ & 1269 / 1252 \\
CrowdPose val & Right elbow & $v{=}2$ & 26.8 & 27.5 & -0.7 [-3.6, 2.2] & 786 / 785 \\
CrowdPose val & Right elbow & $v{=}1$ & 33.4 & 35.3 & -1.9 [-4.5, 0.8] & 1300 / 1278 \\
CrowdPose val & Left knee & $v{=}2$ & 24.4 & 26.8 & -2.5 [-6.6, 1.5] & 552 / 549 \\
CrowdPose val & Left knee & $v{=}1$ & 32.7 & 34.8 & -2.1 [-5.6, 1.4] & 1038 / 997 \\
CrowdPose val & Right knee & $v{=}2$ & 24.7 & 27.4 & -2.7 [-6.5, 1.1] & 553 / 546 \\
CrowdPose val & Right knee & $v{=}1$ & 32.2 & 35.5 & -3.4 [-6.9, 0.1] & 1038 / 1002 \\
OCHuman test & Left elbow & $v{=}2$ & 38.4 & 40.3 & -1.8 [-4.7, 1.0] & 1725 / 1686 \\
OCHuman test & Left elbow & $v{=}1$ & 47.4 & 56.6 & -9.2 [-14.8, -3.6]$^{*}$ & 511 / 460 \\
OCHuman test & Right elbow & $v{=}2$ & 36.9 & 41.0 & -4.1 [-7.1, -1.1]$^{*}$ & 1748 / 1656 \\
OCHuman test & Right elbow & $v{=}1$ & 51.0 & 59.2 & -8.2 [-13.9, -2.6]$^{*}$ & 478 / 459 \\
OCHuman test & Left knee & $v{=}2$ & 23.8 & 26.3 & -2.5 [-5.1, 0.2] & 1351 / 1296 \\
OCHuman test & Left knee & $v{=}1$ & 37.1 & 36.2 & 0.9 [-7.3, 8.9] & 251 / 224 \\
OCHuman test & Right knee & $v{=}2$ & 25.1 & 27.6 & -2.5 [-5.5, 0.5] & 1356 / 1280 \\
OCHuman test & Right knee & $v{=}1$ & 40.0 & 44.0 & -4.0 [-13.2, 5.1] & 266 / 253 \\
\bottomrule
\end{tabular}
}
\vspace{0.4em}\footnotesize Unit: image, resampled independently per model (each model's image set contains images with at least one valid instance of that joint and layer); last column $n_{\mathrm{img}}$: round5-3-oracle / YOLO11s-oracle. 10,000 resamples, seed 42, 95\% percentile CI.
\end{table}

\begin{table}[t]
\centering
\caption{M2 cross-family joint-angle RMSE deltas ($^{\circ}$, per side) under the detector-free oracle protocol with bootstrap CIs. Each cell is $\Delta=$ transformer model $-$ YOLO-oracle model, so $\Delta<0$ favors the transformer (lower error). $^{*}$ CI excludes 0.}
\label{tab:a12}
\renewcommand{\arraystretch}{1.15}
\resizebox{\columnwidth}{!}{%
\begin{tabular}{lllcccc}
\toprule
\textbf{Dataset} & \textbf{Joint} & \textbf{Layer} & \textbf{B$-$r5-3-o} & \textbf{B$-$YOLO11s-o} & \textbf{L$-$r5-3-o} & \textbf{L$-$YOLO11s-o} \\
\midrule
COCO val & Left elbow & $v{=}2$ & -9.6 [-12.1, -7.2]$^{*}$ & -11.2 [-13.8, -8.7]$^{*}$ & -12.4 [-14.8, -9.9]$^{*}$ & -14.0 [-16.4, -11.5]$^{*}$ \\
COCO val & Left elbow & $v{=}1$ & -10.6 [-18.6, -2.5]$^{*}$ & -8.0 [-15.9, -0.3]$^{*}$ & -11.8 [-20.4, -3.3]$^{*}$ & -9.2 [-17.5, -1.0]$^{*}$ \\
COCO val & Right elbow & $v{=}2$ & -8.8 [-11.4, -6.1]$^{*}$ & -9.1 [-11.6, -6.5]$^{*}$ & -11.8 [-14.1, -9.5]$^{*}$ & -12.1 [-14.4, -9.9]$^{*}$ \\
COCO val & Right elbow & $v{=}1$ & -11.0 [-18.7, -3.1]$^{*}$ & -10.1 [-17.6, -2.6]$^{*}$ & -15.4 [-23.2, -7.9]$^{*}$ & -14.5 [-21.9, -7.3]$^{*}$ \\
COCO val & Left knee & $v{=}2$ & -9.8 [-12.6, -7.0]$^{*}$ & -10.4 [-13.2, -7.6]$^{*}$ & -10.7 [-13.5, -7.9]$^{*}$ & -11.3 [-14.2, -8.5]$^{*}$ \\
COCO val & Left knee & $v{=}1$ & -5.4 [-16.2, 5.5] & -6.7 [-16.4, 2.8] & -13.7 [-23.7, -3.5]$^{*}$ & -15.0 [-23.7, -6.2]$^{*}$ \\
COCO val & Right knee & $v{=}2$ & -9.2 [-11.8, -6.6]$^{*}$ & -10.3 [-13.0, -7.6]$^{*}$ & -11.3 [-13.8, -8.8]$^{*}$ & -12.5 [-15.2, -9.8]$^{*}$ \\
COCO val & Right knee & $v{=}1$ & -5.4 [-15.1, 4.3] & -2.8 [-12.3, 6.5] & -8.1 [-17.8, 1.7] & -5.6 [-14.8, 3.7] \\
CrowdPose val & Left elbow & $v{=}2$ & -14.1 [-16.7, -11.5]$^{*}$ & -15.2 [-17.9, -12.5]$^{*}$ & -14.9 [-17.7, -12.2]$^{*}$ & -16.0 [-18.8, -13.1]$^{*}$ \\
CrowdPose val & Left elbow & $v{=}1$ & -1.3 [-3.8, 1.2] & -4.3 [-6.9, -1.7]$^{*}$ & -3.3 [-5.7, -0.9]$^{*}$ & -6.3 [-8.9, -3.7]$^{*}$ \\
CrowdPose val & Right elbow & $v{=}2$ & -12.0 [-14.6, -9.3]$^{*}$ & -12.7 [-15.4, -10.0]$^{*}$ & -13.8 [-16.3, -11.3]$^{*}$ & -14.6 [-17.1, -12.0]$^{*}$ \\
CrowdPose val & Right elbow & $v{=}1$ & -3.6 [-6.3, -1.0]$^{*}$ & -5.5 [-8.2, -2.9]$^{*}$ & -6.9 [-9.5, -4.4]$^{*}$ & -8.8 [-11.5, -6.2]$^{*}$ \\
CrowdPose val & Left knee & $v{=}2$ & -11.4 [-14.8, -8.0]$^{*}$ & -13.9 [-17.4, -10.2]$^{*}$ & -12.3 [-15.9, -8.9]$^{*}$ & -14.8 [-18.3, -11.1]$^{*}$ \\
CrowdPose val & Left knee & $v{=}1$ & -4.5 [-8.1, -0.9]$^{*}$ & -6.7 [-10.3, -3.1]$^{*}$ & -5.4 [-9.0, -1.7]$^{*}$ & -7.5 [-11.1, -3.8]$^{*}$ \\
CrowdPose val & Right knee & $v{=}2$ & -11.5 [-14.8, -8.1]$^{*}$ & -14.2 [-17.5, -10.8]$^{*}$ & -13.3 [-16.5, -10.1]$^{*}$ & -16.0 [-19.2, -12.6]$^{*}$ \\
CrowdPose val & Right knee & $v{=}1$ & -3.2 [-6.9, 0.4] & -6.6 [-10.5, -2.9]$^{*}$ & -5.8 [-9.7, -2.0]$^{*}$ & -9.2 [-13.1, -5.1]$^{*}$ \\
OCHuman test & Left elbow & $v{=}2$ & -1.3 [-4.2, 1.6] & -3.1 [-5.9, -0.2]$^{*}$ & -4.5 [-7.4, -1.6]$^{*}$ & -6.3 [-9.2, -3.4]$^{*}$ \\
OCHuman test & Left elbow & $v{=}1$ & 7.4 [2.4, 12.5]$^{*}$ & -1.8 [-7.2, 3.5] & 4.6 [-0.4, 9.5] & -4.6 [-10.0, 0.8] \\
OCHuman test & Right elbow & $v{=}2$ & -1.0 [-4.1, 2.2] & -5.1 [-8.1, -2.1]$^{*}$ & -3.5 [-6.6, -0.5]$^{*}$ & -7.6 [-10.7, -4.6]$^{*}$ \\
OCHuman test & Right elbow & $v{=}1$ & 4.9 [-0.4, 10.3] & -3.3 [-8.4, 2.0] & 0.6 [-4.6, 5.9] & -7.6 [-12.9, -2.4]$^{*}$ \\
OCHuman test & Left knee & $v{=}2$ & -2.8 [-5.6, -0.0]$^{*}$ & -5.3 [-8.0, -2.5]$^{*}$ & -5.8 [-8.4, -3.1]$^{*}$ & -8.2 [-11.0, -5.6]$^{*}$ \\
OCHuman test & Left knee & $v{=}1$ & -0.3 [-7.7, 7.4] & 0.6 [-7.1, 8.4] & -6.0 [-13.1, 1.1] & -5.1 [-12.4, 2.2] \\
OCHuman test & Right knee & $v{=}2$ & -2.8 [-5.8, 0.1] & -5.3 [-8.3, -2.3]$^{*}$ & -4.7 [-7.7, -1.8]$^{*}$ & -7.2 [-10.3, -4.1]$^{*}$ \\
OCHuman test & Right knee & $v{=}1$ & -3.0 [-10.7, 4.9] & -7.0 [-15.5, 1.5] & -5.3 [-12.9, 2.5] & -9.3 [-17.4, -0.9]$^{*}$ \\
\bottomrule
\end{tabular}
}
\vspace{0.4em}\footnotesize Unit: image, resampled independently per model (each model's image set contains images with at least one valid instance of that joint and layer). 10,000 resamples, seed 42, 95\% percentile CI.
\end{table}
\begin{table}[t]
\centering
\caption{Paired bootstrap CIs for the $v{=}1{-}v{=}2$ layer difference in M2 joint-angle RMSE ($^{\circ}$, per side) within each model, detection-based protocol. $\Delta=$ RMSE$(v{=}1)-$ RMSE$(v{=}2)$, so $\Delta>0$ means the occluded layer has higher error; $^{*}$ CI excludes 0.}
\label{tab:a13}
\renewcommand{\arraystretch}{1.15}
\resizebox{\columnwidth}{!}{%
\begin{tabular}{lllcccc}
\toprule
\textbf{Model} & \textbf{Dataset} & \textbf{Joint} & \textbf{RMSE $v{=}1$} & \textbf{RMSE $v{=}2$} & \textbf{$\Delta$ RMSE [95\% CI]} & \textbf{$n_{\mathrm{img}}$} \\
\midrule
r5-3 & COCO val & Left elbow & 33.5 & 27.0 & 6.5 [1.2, 11.8]$^{*}$ & 248 / 1720 \\
r5-3 & COCO val & Right elbow & 37.7 & 24.9 & 12.8 [8.0, 17.5]$^{*}$ & 290 / 1732 \\
r5-3 & COCO val & Left knee & 30.8 & 20.3 & 10.5 [3.4, 17.2]$^{*}$ & 180 / 1320 \\
r5-3 & COCO val & Right knee & 27.7 & 20.5 & 7.2 [0.2, 14.2]$^{*}$ & 202 / 1302 \\
r5-3 & CrowdPose val & Left elbow & 31.3 & 27.5 & 3.8 [1.3, 6.4]$^{*}$ & 1319 / 790 \\
r5-3 & CrowdPose val & Right elbow & 31.9 & 24.1 & 7.8 [5.6, 10.1]$^{*}$ & 1343 / 810 \\
r5-3 & CrowdPose val & Left knee & 25.7 & 21.1 & 4.5 [1.4, 7.7]$^{*}$ & 1085 / 575 \\
r5-3 & CrowdPose val & Right knee & 26.2 & 20.0 & 6.2 [3.2, 9.1]$^{*}$ & 1104 / 570 \\
r5-3 & OCHuman test & Left elbow & 45.4 & 35.6 & 9.9 [6.6, 13.1]$^{*}$ & 860 / 1931 \\
r5-3 & OCHuman test & Right elbow & 46.0 & 35.6 & 10.4 [7.0, 13.7]$^{*}$ & 822 / 1935 \\
r5-3 & OCHuman test & Left knee & 30.7 & 24.4 & 6.3 [2.5, 10.0]$^{*}$ & 498 / 1507 \\
r5-3 & OCHuman test & Right knee & 33.3 & 24.2 & 9.1 [4.6, 13.3]$^{*}$ & 525 / 1499 \\
YOLO11s & COCO val & Left elbow & 36.6 & 26.3 & 10.3 [4.1, 16.4]$^{*}$ & 247 / 1727 \\
YOLO11s & COCO val & Right elbow & 34.8 & 24.5 & 10.3 [5.6, 15.1]$^{*}$ & 292 / 1730 \\
YOLO11s & COCO val & Left knee & 32.2 & 20.9 & 11.3 [3.3, 18.8]$^{*}$ & 182 / 1325 \\
YOLO11s & COCO val & Right knee & 30.3 & 21.2 & 9.1 [2.2, 15.9]$^{*}$ & 205 / 1306 \\
YOLO11s & CrowdPose val & Left elbow & 35.0 & 27.3 & 7.7 [5.3, 10.2]$^{*}$ & 1321 / 793 \\
YOLO11s & CrowdPose val & Right elbow & 35.0 & 24.6 & 10.4 [7.8, 12.9]$^{*}$ & 1338 / 811 \\
YOLO11s & CrowdPose val & Left knee & 30.5 & 21.3 & 9.2 [6.0, 12.4]$^{*}$ & 1081 / 575 \\
YOLO11s & CrowdPose val & Right knee & 30.5 & 21.2 & 9.3 [6.5, 12.2]$^{*}$ & 1096 / 571 \\
YOLO11s & OCHuman test & Left elbow & 56.2 & 39.7 & 16.5 [13.1, 19.8]$^{*}$ & 871 / 1951 \\
YOLO11s & OCHuman test & Right elbow & 59.2 & 39.5 & 19.7 [16.3, 23.0]$^{*}$ & 831 / 1947 \\
YOLO11s & OCHuman test & Left knee & 39.5 & 28.0 & 11.4 [7.1, 15.8]$^{*}$ & 499 / 1514 \\
YOLO11s & OCHuman test & Right knee & 39.9 & 28.5 & 11.4 [6.9, 15.8]$^{*}$ & 534 / 1503 \\
\bottomrule
\end{tabular}
}
\vspace{0.4em}\footnotesize Unit: image, paired resampling over the union image set (a single resample index is applied to both layers per replicate); the union set contains images with at least one valid instance of that joint in either layer. Last column $n_{\mathrm{img}}$: $v{=}1$ / $v{=}2$ layer image counts. 10,000 resamples, seed 42, 95\% percentile CI.
\end{table}
\begin{table}[t]
\centering
\caption{Benjamini--Hochberg FDR sensitivity analysis over the core comparison families (round5-3 vs.\ YOLO11s-pose). $\Delta=$ round5-3 $-$ YOLO11s-pose: $\Delta>0$ favors round5-3 for AP (F1) and favors YOLO11s-pose for RMSE and jitter (F2, F3).}
\label{tab:a14}
\renewcommand{\arraystretch}{1.15}
\resizebox{\columnwidth}{!}{%
\begin{tabular}{@{}lllrrr@{}}
\toprule
Family & Comparison & $\Delta$ & 95\% CI & $p$ & $p_{\mathrm{BH}}$ \\
\midrule
\multicolumn{6}{@{}l}{\textbf{F1} \ M1 full AP ($n{=}3$; raw sig 3; BH sig 3)} \\
& COCO val & -0.0104 & [-0.0155, -0.0059] & 0.0000 & 0.0000$^{*}$ \\
& CrowdPose val & 0.0687 & [0.0616, 0.0754] & 0.0000 & 0.0000$^{*}$ \\
& OCHuman test & 0.1396 & [0.1281, 0.1508] & 0.0000 & 0.0000$^{*}$ \\
\multicolumn{6}{@{}l}{\textbf{F2} \ M2 per-cell RMSE ($^{\circ}$) ($n{=}24$; raw sig 12; BH sig 11)} \\
& COCO val \ left elbow \ $v{=}2$ & 0.6782 & [-1.7287, 3.0396] & 0.5828 & 0.8322 \\
& COCO val \ left elbow \ $v{=}1$ & -3.0933 & [-11.0757, 4.7844] & 0.4440 & 0.7611 \\
& COCO val \ right elbow \ $v{=}2$ & 0.4619 & [-1.8765, 2.7743] & 0.7022 & 0.8322 \\
& COCO val \ right elbow \ $v{=}1$ & 2.8784 & [-3.4155, 9.1133] & 0.3736 & 0.6897 \\
& COCO val \ left knee \ $v{=}2$ & -0.5693 & [-3.1142, 1.9829] & 0.6554 & 0.8322 \\
& COCO val \ left knee \ $v{=}1$ & -1.3997 & [-11.5768, 8.8087] & 0.7992 & 0.8719 \\
& COCO val \ right knee \ $v{=}2$ & -0.7424 & [-3.1356, 1.6546] & 0.5548 & 0.8322 \\
& COCO val \ right knee \ $v{=}1$ & -2.6246 & [-12.1749, 6.9116] & 0.6016 & 0.8322 \\
& CrowdPose val \ left elbow \ $v{=}2$ & 0.2177 & [-2.5063, 2.9847] & 0.8768 & 0.9149 \\
& CrowdPose val \ left elbow \ $v{=}1$ & -3.7355 & [-6.0494, -1.4487] & 0.0016 & 0.0055$^{*}$ \\
& CrowdPose val \ right elbow \ $v{=}2$ & -0.5567 & [-3.1789, 2.0759] & 0.6718 & 0.8322 \\
& CrowdPose val \ right elbow \ $v{=}1$ & -3.0991 & [-5.4989, -0.7193] & 0.0086 & 0.0188$^{*}$ \\
& CrowdPose val \ left knee \ $v{=}2$ & -0.2008 & [-3.6422, 3.3487] & 0.9206 & 0.9206 \\
& CrowdPose val \ left knee \ $v{=}1$ & -4.8695 & [-7.6710, -2.0897] & 0.0002 & 0.0016$^{*}$ \\
& CrowdPose val \ right knee \ $v{=}2$ & -1.1955 & [-4.3292, 2.0215] & 0.4630 & 0.7611 \\
& CrowdPose val \ right knee \ $v{=}1$ & -4.3532 & [-7.2435, -1.4274] & 0.0030 & 0.0080$^{*}$ \\
& OCHuman test \ left elbow \ $v{=}2$ & -4.1708 & [-6.5563, -1.7360] & 0.0008 & 0.0038$^{*}$ \\
& OCHuman test \ left elbow \ $v{=}1$ & -10.8007 & [-14.5977, -6.8819] & 0.0000 & 0.0000$^{*}$ \\
& OCHuman test \ right elbow \ $v{=}2$ & -3.9219 & [-6.4280, -1.4694] & 0.0016 & 0.0055$^{*}$ \\
& OCHuman test \ right elbow \ $v{=}1$ & -13.2385 & [-17.1184, -9.2323] & 0.0000 & 0.0000$^{*}$ \\
& OCHuman test \ left knee \ $v{=}2$ & -3.6435 & [-6.1476, -1.1277] & 0.0058 & 0.0139$^{*}$ \\
& OCHuman test \ left knee \ $v{=}1$ & -8.7966 & [-13.8743, -3.5585] & 0.0008 & 0.0038$^{*}$ \\
& OCHuman test \ right knee \ $v{=}2$ & -4.3093 & [-6.8603, -1.7329] & 0.0004 & 0.0024$^{*}$ \\
& OCHuman test \ right knee \ $v{=}1$ & -6.6124 & [-12.3613, -0.7771] & 0.0264 & 0.0528 \\
\multicolumn{6}{@{}l}{\textbf{F3} \ M3 temporal (PoseTrack val) ($n{=}7$; raw sig 0; BH sig 0)} \\
& disp mean & -0.0022 & [-0.0062, 0.0018] & 0.3004 & 1.0000 \\
& jitter mean & -0.0014 & [-0.0075, 0.0047] & 0.6510 & 1.0000 \\
& disp P95 & -0.0051 & [-0.0182, 0.0078] & 0.4478 & 1.0000 \\
& jitter P95 & -0.0050 & [-0.0273, 0.0178] & 0.6816 & 1.0000 \\
& jump rate $\tau{=}0.05$ & -0.0091 & [-0.0326, 0.0147] & 0.4490 & 1.0000 \\
& jump rate $\tau{=}0.10$ & -0.0051 & [-0.0188, 0.0085] & 0.4688 & 1.0000 \\
& jump rate $\tau{=}0.20$ & -0.0032 & [-0.0078, 0.0015] & 0.1828 & 1.0000 \\
\bottomrule
\end{tabular}}
\vspace{0.4em}
\footnotesize $^{*}$ BH-adjusted $p<0.05$ within the family. $p$: two-sided bootstrap ($2\min(P(\Delta\le0),P(\Delta\ge0))$), 10,000 resamples, seed 42; significance by whether the 95\% CI excludes 0. F1: paired image universe (union of GT and predicted images of both models); F2: image-level independent resampling; F3: track-segment-level independent resampling. Raw sig counts CI-excludes-0 comparisons before BH adjustment.
\end{table}
\begin{table}[t]
\centering
\caption{M2 joint-angle MAE and P95 ($^{\circ}$, per side) for the YOLO pair, detection-based protocol, stratified by the vertex joint's GT visibility layer. MAE $=$ mean of $|e|$; P95 $=$ 95th percentile of $|e|$; $e=\theta_{\mathrm{pred}}-\theta_{\mathrm{gt}}$ (RMSE restated for cross-check against Table~\ref{tab:a11} and Table~2).}
\label{tab:a15}
\renewcommand{\arraystretch}{1.15}
\resizebox{\columnwidth}{!}{%
\begin{tabular}{llllllll}
\toprule
\textbf{Dataset} & \textbf{Joint} & \textbf{Layer} & \textbf{MAE r5-3} & \textbf{P95 r5-3} & \textbf{MAE YOLO11s} & \textbf{P95 YOLO11s} & \textbf{$n$} \\
\midrule
COCO val & Left elbow & $v{=}2$ & 15.8 & 57.0 & 15.7 & 54.6 & 3148 / 3166 \\
COCO val & Left elbow & $v{=}1$ & 21.4 & 69.7 & 22.8 & 85.7 & 293 / 297 \\
COCO val & Right elbow & $v{=}2$ & 14.4 & 52.4 & 14.5 & 52.6 & 3125 / 3136 \\
COCO val & Right elbow & $v{=}1$ & 24.4 & 94.5 & 22.6 & 77.6 & 356 / 355 \\
COCO val & Left knee & $v{=}2$ & 11.5 & 40.9 & 12.1 & 41.8 & 2513 / 2523 \\
COCO val & Left knee & $v{=}1$ & 17.0 & 75.3 & 17.5 & 69.0 & 207 / 209 \\
COCO val & Right knee & $v{=}2$ & 11.5 & 40.0 & 12.3 & 41.1 & 2467 / 2474 \\
COCO val & Right knee & $v{=}1$ & 16.1 & 57.5 & 17.7 & 69.6 & 230 / 231 \\
CrowdPose val & Left elbow & $v{=}2$ & 16.2 & 59.5 & 16.3 & 59.8 & 2019 / 2022 \\
CrowdPose val & Left elbow & $v{=}1$ & 19.2 & 71.2 & 21.8 & 87.1 & 2925 / 2896 \\
CrowdPose val & Right elbow & $v{=}2$ & 14.8 & 51.7 & 14.9 & 51.8 & 1983 / 1991 \\
CrowdPose val & Right elbow & $v{=}1$ & 19.2 & 73.3 & 21.4 & 85.0 & 2971 / 2937 \\
CrowdPose val & Left knee & $v{=}2$ & 11.6 & 45.0 & 12.3 & 44.6 & 1538 / 1537 \\
CrowdPose val & Left knee & $v{=}1$ & 14.2 & 57.3 & 18.0 & 70.6 & 2270 / 2234 \\
CrowdPose val & Right knee & $v{=}2$ & 10.9 & 37.4 & 12.0 & 39.3 & 1466 / 1468 \\
CrowdPose val & Right knee & $v{=}1$ & 14.0 & 54.9 & 17.6 & 70.6 & 2286 / 2250 \\
OCHuman test & Left elbow & $v{=}2$ & 22.2 & 86.8 & 25.1 & 96.4 & 2634 / 2675 \\
OCHuman test & Left elbow & $v{=}1$ & 33.6 & 96.8 & 42.7 & 120.0 & 905 / 915 \\
OCHuman test & Right elbow & $v{=}2$ & 22.1 & 85.3 & 25.1 & 98.2 & 2672 / 2712 \\
OCHuman test & Right elbow & $v{=}1$ & 33.8 & 99.7 & 45.3 & 120.8 & 854 / 865 \\
OCHuman test & Left knee & $v{=}2$ & 14.1 & 54.6 & 16.9 & 60.6 & 2130 / 2167 \\
OCHuman test & Left knee & $v{=}1$ & 19.4 & 67.4 & 25.5 & 87.6 & 532 / 532 \\
OCHuman test & Right knee & $v{=}2$ & 13.9 & 53.9 & 16.8 & 66.2 & 2100 / 2126 \\
OCHuman test & Right knee & $v{=}1$ & 19.2 & 83.7 & 24.2 & 104.2 & 558 / 566 \\
\bottomrule
\end{tabular}
}
\vspace{0.4em}\footnotesize Unit: instance. Last column $n$: number of valid instances, r5-3 / YOLO11s-pose.
\end{table}


\begin{thebibliography}{18}
\bibitem{1} T.-Y. Lin \emph{et al.}, ``Microsoft COCO: Common objects in context,'' in \emph{Proc. ECCV}, 2014.
\bibitem{2} S.-H. Zhang \emph{et al.}, ``Pose2Seg: Detection free human instance segmentation,'' in \emph{Proc. CVPR}, 2019; arXiv:1803.10683, 2018.
\bibitem{3} M. Andriluka \emph{et al.}, ``PoseTrack: A benchmark for human pose estimation and tracking,'' in \emph{Proc. CVPR}, 2018; arXiv:1712.09184, 2017.
\bibitem{4} J. Turner, J. Clark, and N. Keshtmand, ``Markerless pose estimation for resistance training technique assessment,'' arXiv:2608.24384, 2026.
\bibitem{5} Panconi \emph{et al.}, ``Deep-learning-based markerless pose estimation systems in gait analysis,'' arXiv:2407.10590, 2024.
\bibitem{6} Pemasiri \emph{et al.}, ``Biomechanically accurate gait analysis,'' arXiv:2603.02499, 2026.
\bibitem{7} J. Li \emph{et al.}, ``CrowdPose: Efficient crowded scenes pose estimation and a new benchmark,'' arXiv:1812.00324, 2018.
\bibitem{8} A. Ghafoor and A. Mahmood, ``Quantification of occlusion handling capability of a 3D human pose estimation framework,'' \emph{IEEE Trans. Multimedia}, arXiv:2203.04113, 2022.
\bibitem{9} I. S\'ar\'andi, T. Linder, K. O. Arras, and B. Leibe, ``How robust is 3D human pose estimation to occlusion?'' in \emph{Proc. IROS Workshop on Robotic Co-workers 4.0}, 2018; arXiv:1808.09316, 2018.
\bibitem{10} Y. Garg, S. Bachu, A. Dutta, R. Lal, S. Bose, C.-K. Ta, M. S. Asif, and A. Roy-Chowdhury, ``VOccl3D: A video benchmark dataset for 3D human pose and shape estimation under real occlusions,'' arXiv:2508.06757, 2025.
\bibitem{11} I. L. A. Mendon\c{c}a, F. M. do Nascimento Filho, G. Badicu, D. P. de Vasconcelos, T. B. Lima, V. Calmon, J. P. de Farias Neto, G. K. Jakubiak, L. Y. S. Maciel, F. J. Aidar, and W. Monteiro da Silva J\'unior, ``Analysis of ankle mobility and dynamic valgus knee in runners with and without patellofemoral pain syndrome,'' \emph{Health Science Reports}, vol. 9, e72051, 2026.
\bibitem{12} C. Lei \emph{et al.}, ``Deep video prior for video consistency and propagation,'' \emph{IEEE Trans. Pattern Anal. Mach. Intell.}, arXiv:2201.11632, 2022.
\bibitem{13} G. Casiez, N. Roussel, and D. Vogel, ``1\texteuro\ Filter: A simple speed-based low-pass filter for noisy input in interactive systems,'' in \emph{Proc. CHI}, 2012.
\bibitem{14} G. Jocher \emph{et al.}, ``Ultralytics YOLO11,'' \url{https://github.com/ultralytics/ultralytics}, 2024.
\bibitem{15} A. Zeng, L. Yang, X. Ju, J. Li, J. Wang, and Q. Xu, ``SmoothNet: A plug-and-play network for refining human pose estimation in videos,'' in \emph{Proc. ECCV}, 2022; arXiv:2112.13715, 2022.
\bibitem{16} M. Andriluka, L. Pishchulin, P. Gehler, and B. Schiele, ``2D human pose estimation: New benchmark and state of the art analysis,'' in \emph{Proc. CVPR}, 2014.
\bibitem{17} Y. Xu, J. Zhang, Q. Zhang, and D. Tao, ``ViTPose: Simple vision transformer baselines for human pose estimation,'' in \emph{Proc. NeurIPS}, 2022; arXiv:2204.12484, 2022.
\bibitem{18} A. Kuzdeuov, D. Taratynova, A. Tleuliyev, and H. A. Varol, ``OpenThermalPose: An open-source annotated thermal human pose dataset and initial YOLOv8-pose baselines,'' Authorea Preprints, 2024. \url{https://github.com/Kuzdeuov/OpenThermalPose}
\end{thebibliography}
\end{document}